\documentclass{article}
\usepackage[preprint]{corl_2026} 
\usepackage[utf8]{inputenc} 
\usepackage[T1]{fontenc}    
\usepackage{hyperref}       
\usepackage{url}            
\usepackage{booktabs}      
\usepackage{amsfonts}       
\usepackage{nicefrac}       
\usepackage{microtype}  
\usepackage{array}    
\usepackage[table]{xcolor}
\usepackage{xspace}
\usepackage{subcaption}
\usepackage{multirow}
\usepackage{pifont}
\usepackage{wrapfig}
\usepackage{graphicx}
\usepackage{enumitem}
\definecolor{sharedrow}{RGB}{238,244,252}

\newcommand{\ours}{DeVA\xspace}
\title{\ours: Decoupled Video-Action Model with physical guidance for robot policy learning}

\author{
\makebox[0pt][c]{
Mengqi Zhang\textsuperscript{1} \quad
Sahil Khose\textsuperscript{1} \quad
Simar Kareer\textsuperscript{2} \quad
Yuchen Song\textsuperscript{1} \quad
Unnat Jain\textsuperscript{1} \quad
Judy Hoffman\textsuperscript{1}}\\
\makebox[0pt][c]{
\textsuperscript{1}University of California, Irvine
\quad
\textsuperscript{2}Georgia Institute of Technology}
}

\begin{document}
\hypersetup{
  pdftitle={DeVA: Decoupled Video-Action Model with physical guidance for robot policy learning},
  pdfauthor={Mengqi Zhang, Sahil Khose, Simar Kareer, Yuchen Song, Unnat Jain, Judy Hoffman},
  pdfsubject={Preprint}
}

\makeatletter
\let\@oldmaketitle\@maketitle
\renewcommand{\@maketitle}{\@oldmaketitle
    \centering
    \vspace{-2em}
    \begin{figure}[htbp]
    \centering
    \includegraphics[width=\textwidth]{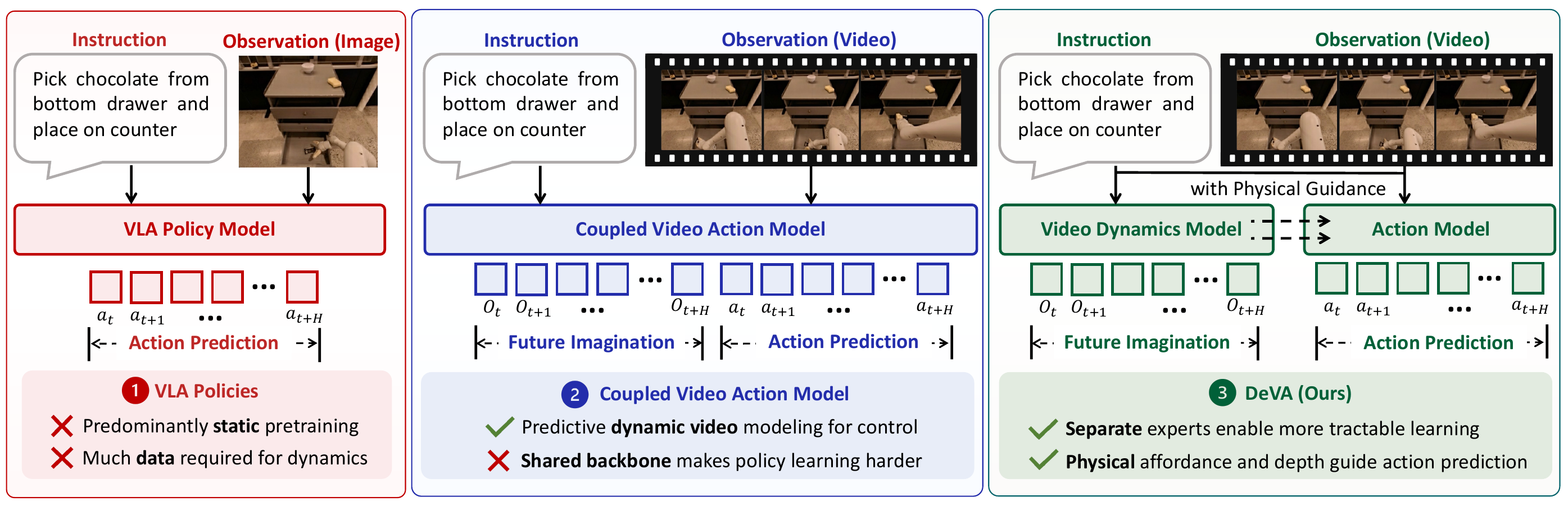}
    \caption{
    \textbf{Overview of \ours compared with VLA and coupled VAM}. VLA directly maps image-language inputs to actions without explicit dynamics modeling, while coupled video-action models jointly process video and action tokens in a shared representation. \ours decouples video dynamics modeling from action prediction with physical guidance, enabling efficient robot policy learning.
    }
    \vspace{-2em}
    \label{fig:teaser}
    \end{figure}
\bigskip}
\maketitle

\begin{abstract}
    Generalizable robot manipulation requires policies that can anticipate how visual scenes evolve while executing language instructions. While recent Vision-Language-Action models benefit from large-scale pretraining, their predominantly static pretraining objectives provide limited supervision for physical dynamics and temporal causality, leaving control-relevant knowledge to be learned from downstream robot demonstrations. Video generative models offer a promising foundation by encoding rich spatiotemporal priors through future predictions. However, existing Video-Action Models either couple video and action prediction in a shared backbone, making policy adaptation harder to optimize, or under-utilize video information when guiding the action branch. In this work, we introduce \textbf{\ours}, a \textbf{De}coupled \textbf{V}ideo-\textbf{A}ction model with specialized video and action experts, multi-level feature transfer, and physically salient guidance. DeVA transfers representations from multiple video layers to the action expert, enabling rich information exchange while making policy learning more tractable. It further supervises intermediate video features and the action stream with physically salient guidance (affordance/depth). Experiments on both simulation benchmarks and real-world deployment demonstrate strong performance with limited data, faster convergence than a unified architecture, and clear performance gains from physical guidance. Additional videos, code, and checkpoints are available at: \href{https://deva-model.github.io/}{deva-model.github.io}  
\end{abstract}

\keywords{Video Action Model, Robot Manipulation, Video Generative Models   } 

\section{Introduction}
	
Vision-Language-Action (VLA) models~\cite{kim2025openvla, black2024pi_0, intelligence2025pi0.5, zitkovich2023rt2} have emerged as a leading paradigm for language-conditioned robotic manipulation by adapting vision-language backbones pretrained on web-scale data. This initialization provides strong semantic priors for instruction following and scene understanding. However, the representations are learned predominantly through static image-text objectives with relatively limited temporal supervision and hence provide constrained priors for physical evolution, temporal causality, and the consequences of robot actions. Consequently, much of this control-relevant knowledge is instead acquired during downstream imitation learning, placing substantial demands on task-specific demonstrations and policy optimization. Video generative models provide a complementary source of knowledge: learning to predict future observations requires modeling how scenes, objects, and interactions evolve over time. This has motivated Video-Action Models (VAMs), which adapt pretrained video backbones to jointly learn future visual dynamics alongside robot actions. Despite this promise, effectively transferring video dynamics to action prediction remains an open architectural challenge, particularly when balancing rich cross-modal interaction with tractable policy optimization.

Existing VAMs address this architectural trade-off through two broad directions. Unified formulations~\cite{zhu2025uwm,kim2025cosmospolicy,ye2026dreamzero} model visual futures and robot actions within a shared backbone or latent representation. Although this design facilitates direct cross-modal exchange, it requires a single feature space to simultaneously support visual generation and control prediction, which restricts the ability to learn the modality-specific features necessary for solving the individual tasks. Earlier approaches~\cite{hu2025video} repurpose pretrained video models as predictive representation encoders for downstream policies, whereas concurrent dual-DiT architectures~\cite{pai2025mimicvideo,ma2026dit4dit} allow an action transformer to attend to intermediate video features. Nevertheless, these interfaces typically rely on representations extracted from a selected backbone layer at a fixed denoising stage, leaving the complementary abstractions distributed across the video backbone insufficiently explored. Moreover, whether a unified or dual architecture, prior work primarily acquired control-relevant features implicitly through video and action objectives and did not explore the integration of task-relevant signals such as scene geometry and interaction regions. An important open problem is therefore how to efficiently guide action learning from video pretraining while preserving modality-specific features and focusing on the information from videos most relevant for actions.

To address these challenges, we introduce \ours, a decoupled Video-Action Model that combines specialized video and action experts with structured multi-level feature transfer and physically salient guidance. Rather than requiring a single representation to support both visual generation and control, \ours assigns the two objectives to dedicated transformer branches, preserving modality-specific capacity and making downstream policy optimization more tractable. The decoupling, however, introduces the need for an effective interface between the two experts. \ours therefore routes intermediate representations from multiple levels of the video backbone to corresponding action blocks through layer-wise cross-attention, augmented with learnable bridge tokens that compactly aggregate video dynamics into self-attention. This interface exposes the action expert to comprehensive spatiotemporal representations distributed across the video backbone, building on observations that diffusion features encode distinct abstractions across layers and denoising stages~\cite{luo2023diffusion,kim2025revelio}. To further emphasize physical information relevant to manipulation, \ours applies affordance and relative-depth supervision to intermediate video features. These auxiliary objectives encourage the video backbone to encode task-relevant interaction regions and scene geometry, while the resulting decoder features are injected into the action stream as explicit physical guidance. Together, these components provide an effective framework that combines tractable expert specialization with rich and physically informed video-to-action learning.

We evaluate \ours on three simulation benchmarks, including RoboCasa~\cite{robocasa2024}, LIBERO~\cite{liu2023libero}, LIBERO-plus~\cite{fei2025liberoplus}, and on real-world bimanual manipulation tasks. Across these settings, \ours achieves stronger or competitive performance against conventional diffusion policy, finetuned VLA models, and recent Video-Action Models, despite using limited task-specific demonstrations. Under a matched optimization budget, \ours converges faster and outperforms its unified counterpart. Controlled comparisons further show improvements from affordance and depth guidance, supporting the benefits of the decoupled architecture and physically salient supervision.
\vspace{-1em}

\section{Related Work}
\vspace{-1em}
\paragraph{Vision-Language-Action Models.} 
Vision-Language-Action (VLA) models adapt pretrained vision-language representations for language-conditioned robot control. Early generalist policies established the benefits of scaling robot data and transferring web knowledge ~\cite{brohan2022rt,driess2023palm,zitkovich2023rt2,bousmalis2023robocat}. Subsequent models expanded open-source pretraining, cross-embodiment learning, and continuous action generation~\cite{kim2025openvla,black2024pi_0,octo_2023,li2024cogact,liu2025rdt,bjorck2025grootn1}. Recent work further improves action tokenization and adaptation~\cite{intelligence2025pi0.5,pertsch2025pifast,kim2025openvla-oft}, spatial and temporal grounding~\cite{zhen20243d,qu2025spatialvla,zheng2025tracevla}, and unified policy learning~\cite{bu2025univla,li2025cogvla}. These models inherit strong semantic priors, but their pretraining remains predominantly oriented toward static visual understanding. \ours instead investigates video-generative pretraining as a complementary source of predictive dynamics for policy learning.
\vspace{-1em}

\paragraph{Video-based Action Models.} 
Video prediction has been used for robot control through generated subgoals, inverse dynamics, and predictive representations~\cite{hu2025video,du2023learning,black2024zero,bharadhwaj2025gen2act,jang2025dreamgen}. Related approaches learn latent actions or use future prediction as an auxiliary policy objective~\cite{ye2025lapa,wu2024unleashing,zheng2025flare}. More recent Video-Action Models (VAMs) model visual futures and actions more directly. Unified formulations place both modalities in a shared backbone or latent space~\cite{zhu2025uwm,kim2025cosmospolicy,ye2026dreamzero,li2025uva}, whereas modular formulations connect separate video and action networks through predicted features or attention~\cite{pai2025mimicvideo,ma2026dit4dit,liang2025videopolicy,routray2025vipra,yuan2026fast}. \ours follows the modular direction but combines specialized experts with multi-level transfer across multiple video layers, providing richer video-to-action interaction without requiring a shared representation.
\vspace{-1em}

\paragraph{Physical Grounding for Manipulation.}
Affordance and geometry provide spatial structure for identifying where and how a robot should interact with its environment. Prior work learns actionable regions, object motion, or task-conditioned affordances from images, 3D observations, and human videos~\cite{mo2021where2act,deng20213d,bahl2023affordances,yuan2025general,yuan2025robopoint,huang2025rekep}. Affordance representations have also been incorporated into hierarchical and diffusion-based policies to improve transfer and demonstration efficiency~\cite{nasiriany2025rt,rana2025learning,wu2025afforddp,tang2025uad}. Complementary approaches use depth, point clouds, or 3D feature fields for spatially grounded action prediction~\cite{shridhar2023perceiver,goyal2023rvt,gervet2023act3d,Ze2024DP3,jia2024lift3d,yang2024depth,chen2025video,depth_anything_v1,fan2026aim}. Rather than treating these cues only as policy inputs or external planning outputs, \ours uses affordance and relative-depth supervision to shape intermediate video representations and directly guide the action expert.
\vspace{-1em}
\section{Method}
\vspace{-1em}

In this section, we present \ours, a decoupled Video-Action Model that adapts the video generative backbone for robot policy learning through specialized video and action experts, multi-level feature interaction, and physically salient guidance. Given the current observation $O_t$ and instruction $T$, the video expert predicts future observations \(\{O_{t+1}, \dots, O_{t+h}\}\) that capture plausible scene evolution over a horizon $h$. In parallel, the action expert predicts the corresponding action sequence \(\{A_{t}, \dots, A_{t+h-1}\}\). Together, the two experts model \(P(\{O_{t+1}, \dots, O_{t+h}\}, \{A_t, \dots, A_{t+h-1}\}\mid O_t, T)\), linking predictive visual dynamics to executable control while maintaining specialized representations. \ours injects intermediate representations from multiple levels of the video backbone to the action expert. It further applies affordance and depth supervision to the video features and uses the resulting physical features to guide action prediction. We describe the architecture in Sec.~\ref {sec3:model_arch}, the physical guidance mechanism in Sec.~\ref{sec3:guidance}, and the training and inference pipeline in Sec.~\ref{sec3:pipeline}.
\vspace{-0.5em}

\subsection{Decoupled Video-Action Architecture with Multi-Level Feature Interaction}
\label{sec3:model_arch}
\begin{figure*}[t]
    \centering
    \includegraphics[width=0.95\textwidth]{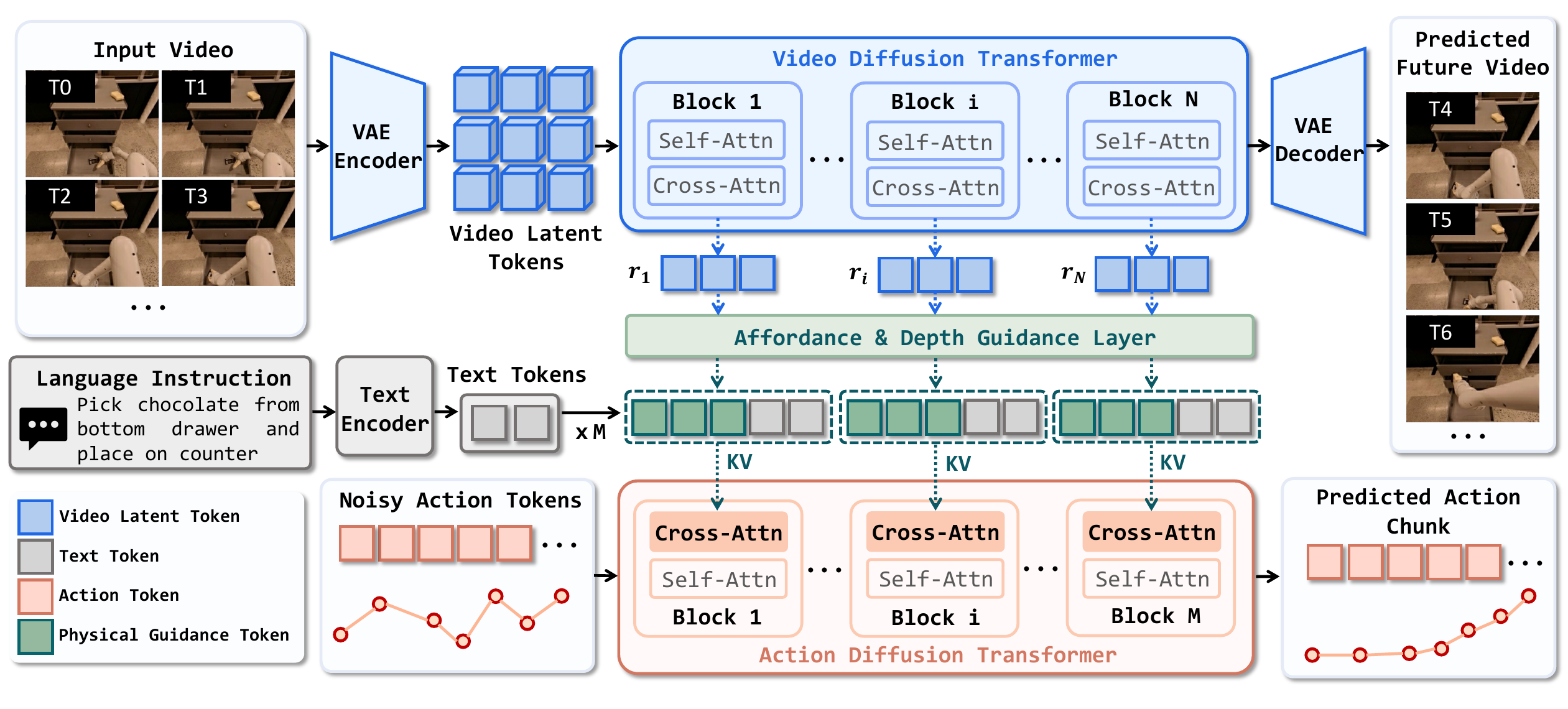}
    \caption{
    \textbf{Overview.} DeVA decouples video dynamics modeling and action prediction. The action stream attends to layer-wise video features through cross-attention to produce robot action chunks.}
    \label{fig:dawn_architecture}
    \vspace{-1.5em}
\end{figure*}
\vspace{-0.5em}
Unlike some prior VAMs that model video and action prediction within a shared representation space~\citep{kim2025cosmospolicy,ye2026dreamzero}, \ours in contrast assigns visual dynamics modeling and action prediction to specialized transformer experts. This separation maintains a dedicated pathway for predictive visual modeling while providing action-specific capacity for robot control. To remain connected, the two experts leverage multi-level feature interaction, implemented using layer-wise cross-attention and learnable bridge tokens.
\vspace{-1em}

\paragraph{Decoupled Architecture.}

\ours consists of a video expert for future observation modeling and an action expert for robot action prediction (Fig.~\ref{fig:dawn_architecture}). The video expert is initialized from Cosmos-Predict2~\cite{nvidia2025cosmospredict2}, a latent video diffusion transformer. Its spatiotemporal VAE encodes video observations into compact latent tokens, and the video transformer models the evolution conditioned on the current frame and language instruction. The action expert follows a similar DiT-style architecture in the action domain and predicts the action trajectory aligned with the corresponding predicted future. The two experts maintain separate parameters and token representations for their respective objectives. This avoids requiring visual dynamics and robot actions to share a single feature space. At the same time, the experts are jointly optimized using separate generative objectives, allowing the action policy to benefit from the video dynamics while retaining its specialized control representation.
\vspace{-1em}

\paragraph{Multi-Level Feature Interaction.}
Although separate experts preserve modality-specific capacity, fully isolating them would prevent the action expert from accessing the predictive dynamics learned by the video backbone. Effective decoupling therefore requires structured interaction between the two experts. Moreover, diffusion models encode complementary representations across network layers and denoising stages~\cite{luo2023diffusion,kim2025revelio}. \ours transfers intermediate video representations from multiple levels to corresponding action blocks through layer-wise cross-attention. Learnable bridge tokens provide an additional compact interface by aggregating video context and introducing it into the action stream through self-attention. Together, these mechanisms expose the action expert to complementary predictive representations distributed across the video backbone while preserving its specialized control capacity.
\vspace{-0.8em}

\subsection{Physically Salient Guidance}
\vspace{-0.5em}
\label{sec3:guidance}
The multi-level interaction interface provides the action expert with predictive video representations. However, these representations are optimized primarily for future prediction and may not explicitly emphasize the interaction regions and geometric structure needed for successful control. \ours therefore applies task-conditioned affordance and relative-depth supervision to intermediate video features. The resulting decoder features are also injected into the action expert as physical guidance. This mechanism both shapes the video representations and directly conditions action prediction.
\vspace{-1em}

\begin{figure}[t]
    \centering
    \includegraphics[width=\linewidth]{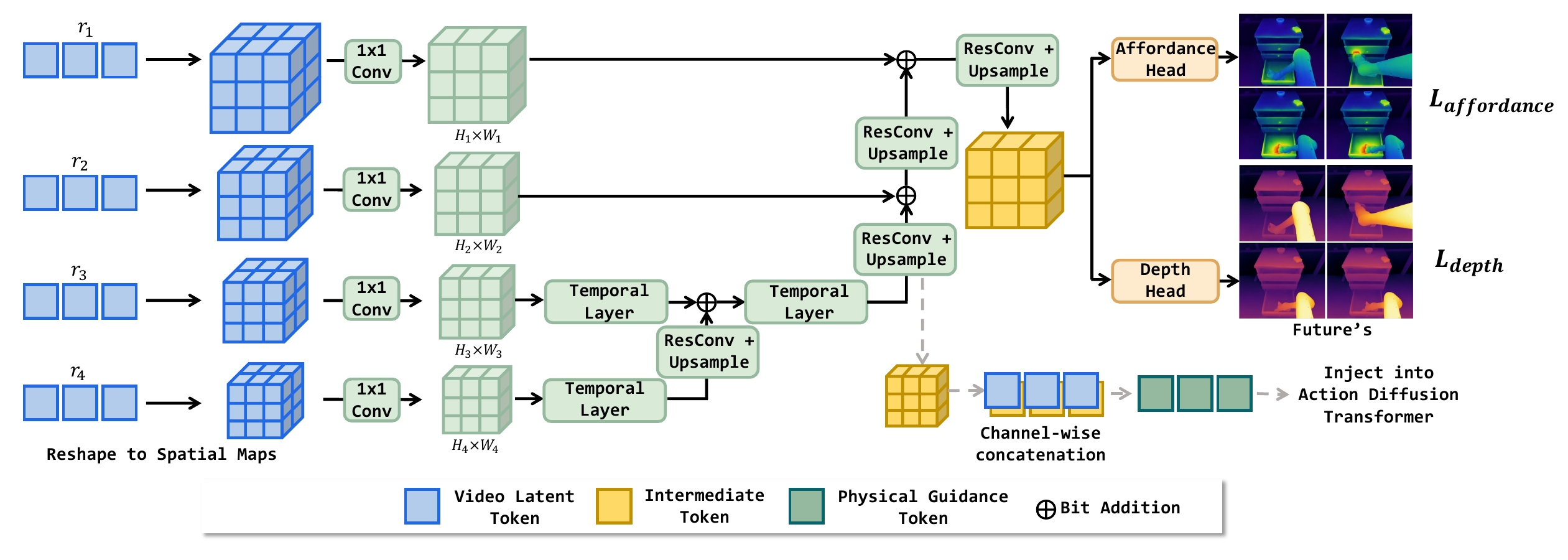}
    \caption{\textbf{Physical guidance module.} The module aggregates multi-level video features and decodes them into affordance and depth maps, which are then converted into physically grounded tokens and injected into the action transformer head to support action prediction.}
    \vspace{-1.5em}
    \label{fig:guidance_head}
\end{figure}

\paragraph{Affordance \& Depth Decoding.}
We repurpose intermediate video features as dense representations for manipulation perception. Concretely, \ours uniformly samples features from the video backbone layers and feeds them into lightweight DPT-style decoders with interleaved temporal-attention layers, as illustrated in Fig.~\ref{fig:guidance_head}. The decoders predict two complementary physical signals: task-conditioned affordance maps and relative-depth maps. We define the affordance map over the image plane as
\[
\mathcal{A}_t(u,v)
=
P\bigl(p_t^{ee}=(u,v)\mid O_t,T\bigr),
\]
where \(p_t^{ee}\) denotes the image-plane location of the end effector and each pixel represents its likelihood as an interaction location under instruction \(T\). Language features are incorporated into the affordance decoder through FiLM layers. In simulation, we construct affordance targets from ground-truth contact locations or projected end-effector positions and convert them into smooth heatmaps using Gaussian kernels. For real-world data, where such annotations are unavailable, we generate pseudo-labels using an off-the-shelf affordance model~\cite{tang2025uad}. We similarly obtain relative monocular depth estimations with a depth prediction model~\cite{chen2025video}. Additional details are provided in the Appendix.
\vspace{-1em}

\paragraph{Physical Guidance Injection.}
Beyond providing auxiliary supervision, the decoded physical features directly guide the action expert. We extract features from the final spatiotemporal blocks of the decoders, project them to the spatial resolution of the video features, and flatten them into tokens. These features are concatenated with the multi-level video representations along the channel dimension and supplied as additional keys and values to the action cross-attention layers. The action expert can therefore attend to both predictive video dynamics and explicit cues about where and how the end effector should act. In this way, physically salient guidance both emphasizes control-relevant structure in the video backbone and makes this structure directly available to action prediction.

\vspace{-0.8em}

\subsection{Training and Inference}
\label{sec3:pipeline}
\vspace{-0.5em}
The separation between the video and action experts supports a two-stage training procedure. In Stage 1, we warm up the video expert and physical decoders for 10K steps, jointly learning future visual dynamics, affordance, and depth. In Stage 2, we introduce the action expert and jointly optimize video and action prediction. The physical decoders are frozen during this stage, but continue to supervise the video features and provide guidance to the action expert.
\vspace{-1em}

\paragraph{Stage 1: Video and Decoder Warmup.}
The video expert is trained in latent space using the standard EDM denoising objective \(\mathcal{L}_{\mathrm{video}}\). We additionally supervise the affordance and depth decoders to encourage the intermediate video features to encode physically salient structure. The affordance decoder is trained using a noise-weighted pixel-wise binary cross-entropy loss:
\[
\mathcal{L}_{\mathrm{aff}}
=
\mathbb{E}_{t}
\left[
w(t)\,
\ell_{\mathrm{BCE}}\big(\hat{\mathcal{A}}_t,\mathcal{A}_t\big)
\right],
\qquad
w(t)=\frac{1}{(1+\sigma_t)^2},
\]
where \(t\) denotes the diffusion noise step and \(\sigma_t\) is the corresponding noise level. The loss is averaged over spatial locations and includes positive-class reweighting to address the foreground-background imbalance in the affordance maps. The depth decoder is trained using a noise-weighted masked reconstruction loss:
\[
\mathcal{L}_{\mathrm{depth}}
=
\mathbb{E}_{t}
\left[
w(t)
\left(
\ell_{1}^{\mathrm{mask}}\big(\hat{D}_t, D_t\big)
+
\lambda_{\mathrm{grad}}
\ell_{\mathrm{grad}}^{\mathrm{mask}}\big(\hat{D}_t, D_t\big)
\right)
\right],
\]
where \(\ell_{1}^{\mathrm{mask}}\) is computed only over valid depth pixels, while \(\ell_{\mathrm{grad}}^{\mathrm{mask}}\) encourages local geometric consistency. The overall objective in Stage 1 is:
\[
\mathcal{L}_{\mathrm{v}}
=
\mathcal{L}_{\mathrm{video}}
+
\lambda_{\mathrm{aff}} \mathcal{L}_{\mathrm{aff}}
+
\lambda_{\mathrm{depth}} \mathcal{L}_{\mathrm{depth}}.
\]
\vspace{-2.5em}
\paragraph{Stage 2: Joint Video-Action Training.}
In Stage 2, we introduce the action expert and jointly optimize it with the video expert. The parameters of the physical decoders are frozen to maintain a stable physical prediction interface. However, the affordance and depth predictions remain active, allowing their gradients to pass through the fixed decoders and continue shaping the intermediate video features. The decoded physical features are also retained as direct guidance for the action expert. The action expert is trained using a flow-matching objective \(\mathcal{L}_{\mathrm{act}}\) conditioned on the current observation, language instruction, multi-level video representations, and decoded physical guidance. The complete Stage 2 objective is:
\[
\mathcal{L}
=
\mathcal{L}_{\mathrm{v}}
+
\mathcal{L}_{\mathrm{act}}.
\]
\vspace{-3em}

\paragraph{Inference.}
At inference time, the future video latents and action trajectory are initialized from Gaussian noise and sampled jointly through their respective generative processes. The video expert predicts a future visual rollout, while the action expert generates the corresponding action chunk. The lightweight physical decoders remain active to provide affordance- and depth-aware guidance.

\vspace{-1em}
\section{Experiment}

\begin{figure}[t]
    \centering
    \resizebox{0.98\linewidth}{!}{%
    \begin{minipage}[t]{0.625\textwidth}
        \centering
        \vspace{0pt}
        \includegraphics[width=\linewidth]{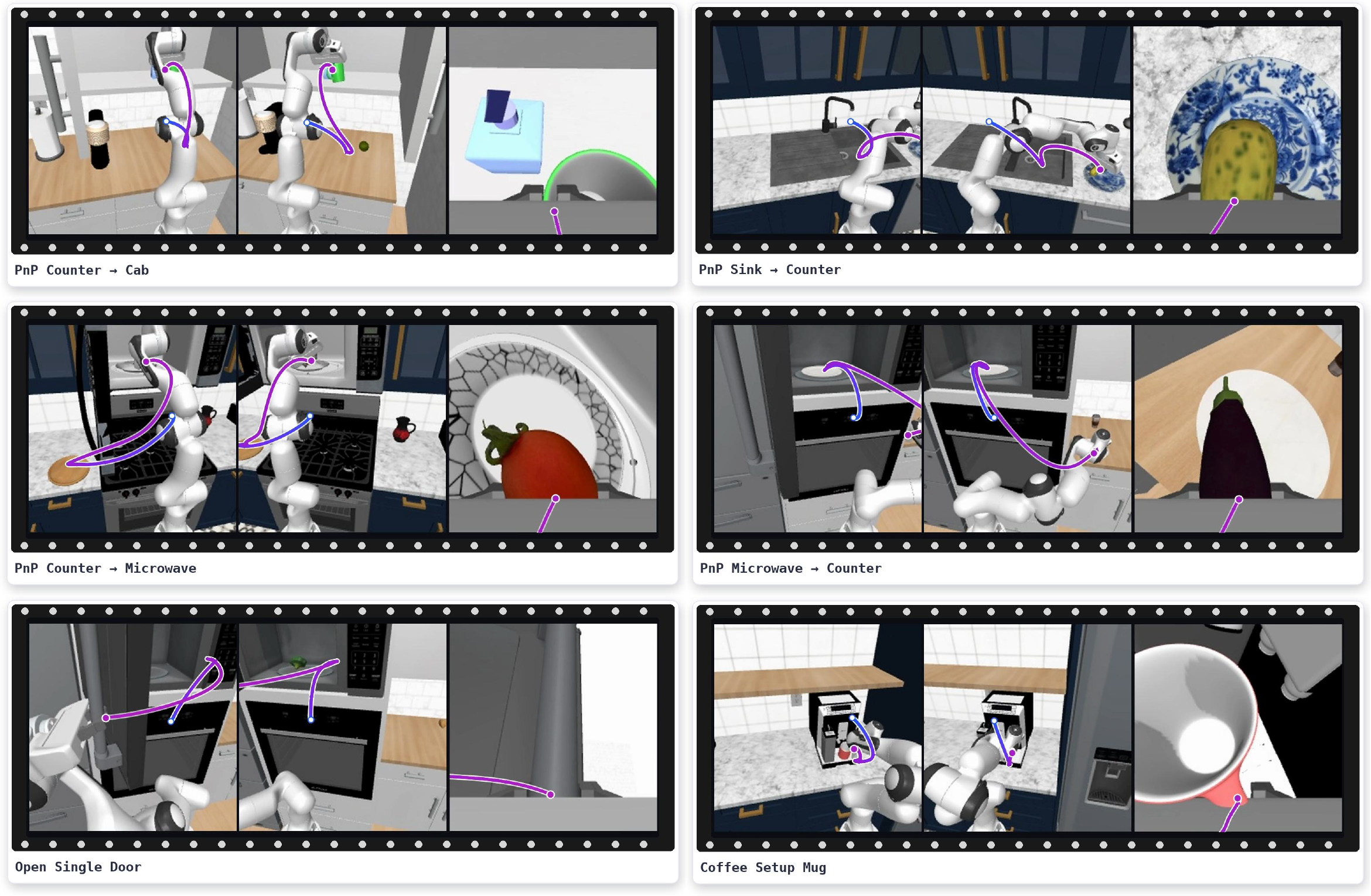}
        \vspace{-1.3em}
        \captionof{figure}{RoboCasa task rollout examples.}
        \label{fig:robocasa_qual}
    \end{minipage}
    \hfill
    \begin{minipage}[t]{0.45\textwidth}
        \centering
        \vspace{0pt}
        \scriptsize
\setlength{\tabcolsep}{3pt}
\renewcommand{\arraystretch}{1.05}
\resizebox{\linewidth}{!}{%
\begin{tabular}{lcc}
\toprule
\textbf{Success Rate} & \textbf{\# Demos / Task} & \textbf{Avg.} \\
\midrule
GR00T-N1~\cite{bjorck2025grootn1} & 300 & 49.6 \\
DP-VLA~\cite{han2024dpvla} & 3000 & 57.3 \\
GR00T-N1 + DreamGen~\cite{jang2025dreamgen} & 300 (+10K syn.) & 57.6 \\
GR00T-N1 + DUST~\cite{won2025dust} & 300 & 58.5 \\
$\pi_0$~\cite{black2024pi_0} & 300 & 62.5 \\
GR00T-N1.5~\cite{bjorck2025grootn1} & 300 & 64.1 \\
FLARE~\cite{zheng2025flare} & 300 & 66.4 \\
GR00T-N1.5 + HAMLET~\cite{koo2025hamlet} & 300 & 66.4 \\
\midrule
\multicolumn{3}{c}{\textbf{Video-Action Models}} \\
\midrule
Video Policy~\cite{liang2025videopolicy} & 300 & 66.0 \\
UVA~\cite{li2025uva} & 50 & 50.0 \\
UWM~\cite{zhu2025uwm} & 1000 & 60.8 \\
Cosmos-Policy~\cite{kim2025cosmospolicy} & 50 & \underline{67.1} \\
\textbf{Ours} & 50 & \textbf{72.0} \\
\bottomrule
\end{tabular}%
}
        \captionof{table}{RoboCasa benchmark comparison.}
        \label{tab:robocasa}
    \end{minipage}%
    }
    \vspace{-2em}
\end{figure}
\vspace{-0.8em}

In this section, we evaluate and explore the potential of \ours from three perspectives: benchmark comparison over state-of-the-art policy models; data efficiency and speedup based on our two-flow architecture; scalable explorations over different axes.
\vspace{-1em}
\subsection{Benchmarks}
\vspace{-0.8em}

\begin{wrapfigure}{r}{0.52\textwidth}
    \vspace{-2em}
    \centering
    \small

    \begin{minipage}[t]{\linewidth}
        \centering
        \vspace{0pt}
        \small
\setlength{\tabcolsep}{4pt}
\renewcommand{\arraystretch}{1.08}
\resizebox{\linewidth}{!}{%
\begin{tabular}{lcccccc}
\toprule
\textbf{Success Rate} & \textbf{Spatial} & \textbf{Object} & \textbf{Goal} & \textbf{Long} & \textbf{Short-Avg} & \textbf{Avg.} \\
\midrule
Diffusion Policy~\cite{chi2025diffusionpolicy} & 78.3 & 92.5 & 68.3 & 50.5 & 79.7 & 72.4 \\
$\pi_0$-fast~\cite{pertsch2025pifast} & 96.4 & 96.8 & 88.6 & 60.2 & 93.9 & 85.5 \\
$\pi_{0.5}$~\cite{intelligence2025pi0.5}  & \underline{98.8} & 98.2 & 98.0 & 92.4 & 98.3 & 96.9 \\
OpenVLA-OFT~\cite{kim2025openvla-oft}  & 97.6 & 98.4 & 97.9 & 94.5 & 98.0 & 97.1 \\
CogVLA~\cite{li2025cogvla} & 98.6 & 98.8 & 96.6 & 95.4 & 98.0 & 97.4 \\
\midrule
\multicolumn{7}{c}{\textbf{Video-Action Models}} \\
\midrule
UVA~\cite{li2025uva} & - & - & - & 90.0 & - & - \\
UWM~\cite{zhu2025uwm} & - & - & - & 79.0 & - & - \\
ViPRA~\cite{routray2025vipra} & - & - & - & 79.0 & - & -\\
Mimic-Video~\cite{pai2025mimicvideo} & 94.2 & 96.8 & 90.6 & - & 93.9 & - \\
Cosmos-Policy~\cite{kim2025cosmospolicy} & 98.1 & \textbf{100.0} & 98.2 & \underline{97.6} & 98.8 & 98.5 \\
DiT4DiT~\cite{ma2026dit4dit}& 98.4 & \underline{99.6} &\underline{98.6} &\underline{97.6} &\underline{98.9}&\underline{98.6}\\
\textbf{Ours} & \textbf{99.2} & \underline{99.6} & \textbf{98.8} & \textbf{98.4} & \textbf{99.2} & \textbf{99.0} \\
\bottomrule
\end{tabular}%
}
        \vspace{-0.5em}
        \caption*{\small (a) LIBERO benchmark comparison.}
    \end{minipage}

    \begin{minipage}[t]{0.49\linewidth}
        \centering
        \vspace{0pt}
        \small
\setlength{\tabcolsep}{5pt}
\renewcommand{\arraystretch}{1.08}
\resizebox{\linewidth}{!}{%
\begin{tabular}{lc}
\toprule
\textbf{Success Rate} & \textbf{Avg.} \\
\midrule
OpenVLA~\cite{kim2025openvla} & 15.6 \\
WorldVLA~\cite{cen2025WorldVLA} & 25.0 \\
NORA~\cite{hung2025nora} & 39.0 \\
UniVLA~\cite{bu2025univla} & 42.9 \\
$\pi_0$~\cite{black2024pi_0} & 53.6 \\
OpenVLA-OFT\_w~\cite{kim2025openvla-oft} & 55.8 \\
$\pi_0$-fast~\cite{pertsch2025pifast} & 61.6 \\
OpenVLA-OFT\_m~\cite{kim2025openvla-oft} & 67.9 \\
RIPT-VLA~\cite{tan2025ript-vla} & 68.4 \\
OpenVLA-OFT~\cite{kim2025openvla-oft} & \underline{69.6} \\
\textbf{Ours} & \textbf{80.8} \\
\bottomrule
\end{tabular}%
}
        \vspace{-0.5em}
        \caption*{\small (b) LIBERO-Plus.}
    \end{minipage}\hfill
    \begin{minipage}[t]{0.47\linewidth}
        \centering
        \vspace{-3pt}
        \includegraphics[width=\linewidth]{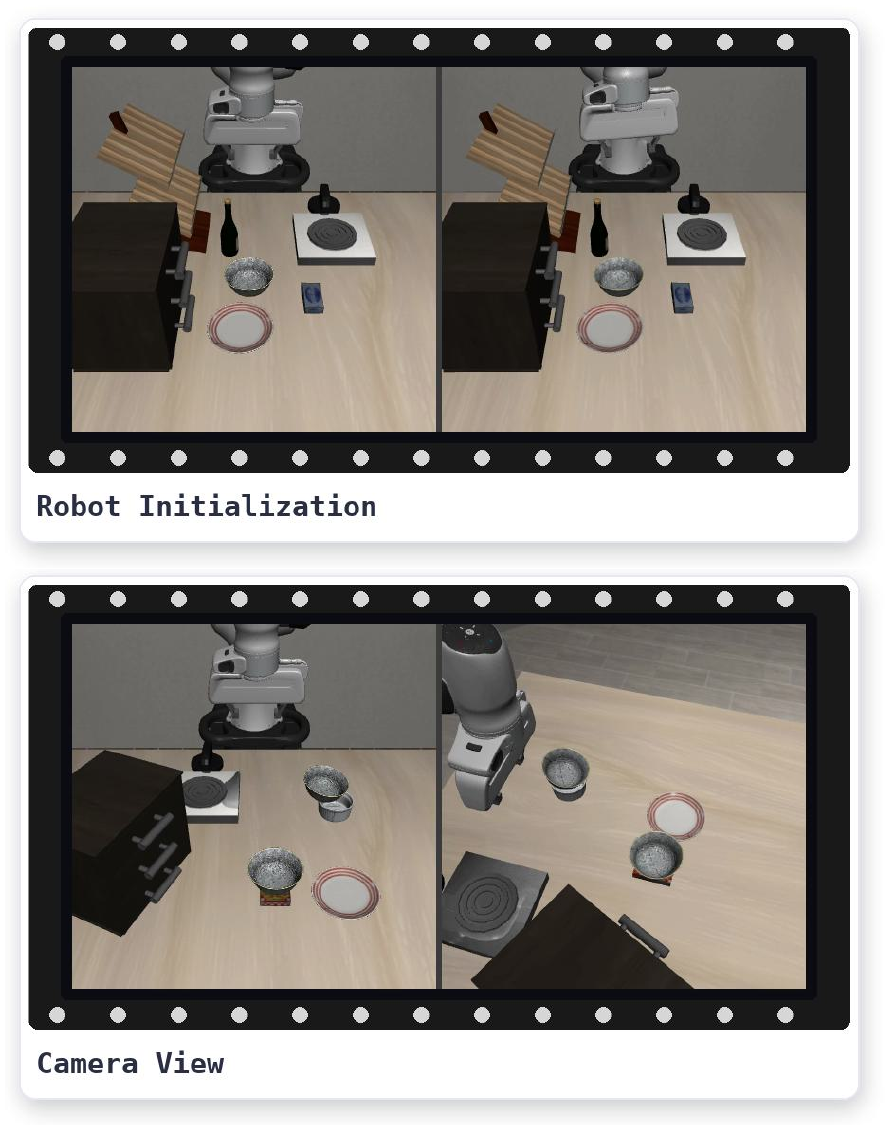}
        \vspace{-2em}
        \caption*{\small (c) Perturbation.}
    \end{minipage}

    \vspace{-0.5em}
    \caption{Simulation benchmark comparisons on LIBERO and LIBERO-Plus. The robustness examples are on "Robot Init" and "Camera View" axes.}
    \label{fig:libero_results}
    \vspace{-3em}
\end{wrapfigure}

\paragraph{Simulation Benchmarks.}We evaluate \ours on three complementary simulation benchmarks covering multitask manipulation, diverse environments, and robustness under controlled shifts:

\vspace{-0.8em}
\begin{itemize}[leftmargin=1.2em, itemsep=0.2em, topsep=0.2em]
    \item \textbf{RoboCasa~\cite{robocasa2024}.} A large-scale simulation benchmark featuring realistic kitchen environments and diverse everyday manipulation tasks, including pick-and-place and appliance operation. 
    \vspace{-0.1em}
    \item \textbf{LIBERO~\cite{liu2023libero}.} A tabletop manipulation benchmark organized into task suites that vary in spatial layouts, object identities, goal specifications, and task horizons.
    \vspace{-0.1em}
    \item \textbf{LIBERO-Plus~\cite{fei2025liberoplus}.} A robustness benchmark extending LIBERO with controlled perturbations across seven dimensions: object layout, camera viewpoint, robot initial state, language instruction, lighting condition, background texture, and sensor noise.
\end{itemize}
\vspace{-0.4em}
Together, these benchmarks evaluate policy performance across controlled tabletop tasks, visually diverse household environments, and systematic distribution shifts.
\vspace{-1em}

\paragraph{Real World Manipulation.} 
Our real-world bimanual manipulation platform consists of two I2RT YAM Ultra robotic arms~\cite{i2rt_yam_arm_series}. Each arm provides 6-DoF motion and a 1-DoF parallel gripper, resulting in a total of 14-DoF. The perception system uses three RGB-D cameras: one wrist-mounted camera on each arm for close-range observations and one external head-view camera for global scene context. Expert demonstrations are collected using a Meta Quest Pro teleoperation interface based on OpenTeleVision~\cite{cheng2025open}. We evaluate \ours on three bimanual tasks: handing over a marker between arms, lifting a pot, and picking up bottles of diverse appearances. These tasks test different forms of bimanual control, including inter-arm handoff, coordinated object handling, and object variations.
\vspace{-1em}

\subsection{Baselines}
\vspace{-0.8em}
Depending on the benchmark, we compare \ours with baselines from three model families. First, conventional task-specific policies are represented by Diffusion Policy~\cite{chi2025diffusionpolicy}. Second, VLA models include OpenVLA and its variants~\cite{kim2025openvla}, CogVLA~\cite{li2025cogvla}, the \(\pi\)-series~\cite{black2024pi_0,intelligence2025pi0.5,pertsch2025pifast}, and GR00T-series~\cite{bjorck2025grootn1}. Third, video-action baselines include Video Policy~\cite{liang2025videopolicy}, UVA~\cite{li2025uva}, UWM~\cite{zhu2025uwm}, ViPRA~\cite{routray2025vipra}, Mimic-Video~\cite{pai2025mimicvideo}, Cosmos Policy~\cite{kim2025cosmospolicy}, and DiT4DiT~\cite{ma2026dit4dit}.

\vspace{-0.8em}
\subsection{Simulation Results}
\vspace{-0.5em}
\paragraph{RoboCasa.} 
We evaluate \ours on 24 kitchen manipulation tasks. For each task, we conduct 50 evaluation rollouts for three different seeds, resulting in 3,600 rollouts in total. The benchmark covers a range of manipulation challenges. Cabinet and microwave tasks require object retrieval in cluttered or constrained workspaces, while mug placement and stove manipulation demand precise object localization and control, as illustrated in Fig.~\ref{fig:robocasa_qual}. As shown in Table~\ref{tab:robocasa}, \ours achieves a success rate of 72.0\%, outperforming different types of baselines. Importantly, \ours uses only 50 demonstrations per task. Under the same data budget, it improves up to 22.0 points, while also outperforming several methods trained with 300--3,000 demonstrations per task. These results demonstrate strong data efficiency without relying on additional synthetic trajectories.
\vspace{-1em}

\paragraph{LIBERO \& LIBERO-plus.} 
We first evaluate \ours on the four standard LIBERO suites: Spatial, Object, Goal, and Long, comprising 40 tasks in total. We evaluate every task over 50 rollouts using three seeds. We report the average success rate in Fig.~\ref{fig:libero_results}. \ours achieves the highest overall average success rate of 99.0\%. For VAM baselines that report only LIBERO-Long, we restrict the comparison to that suite. Overall, \ours maintains consistently high performance across spatial, object-centric, goal-conditioned, and long-horizon tasks. We further evaluate robustness on LIBERO-Plus, which contains 10,030 variants generated by applying controlled perturbations to the original LIBERO tasks. Following the official protocol, each variant is evaluated with one rollout. \ours achieves an average success rate of 80.8\%, exceeding the strongest baseline shown in the table, OpenVLA-OFT, by 11.2 percentage points. The 18.2-point drop from 99.0\% on standard LIBERO also reveals a remaining gap between in-distribution and robustness.
\begin{wrapfigure}{r}{0.47\textwidth}
    \vspace{-0.5em}
    \centering
    \includegraphics[width=\linewidth]{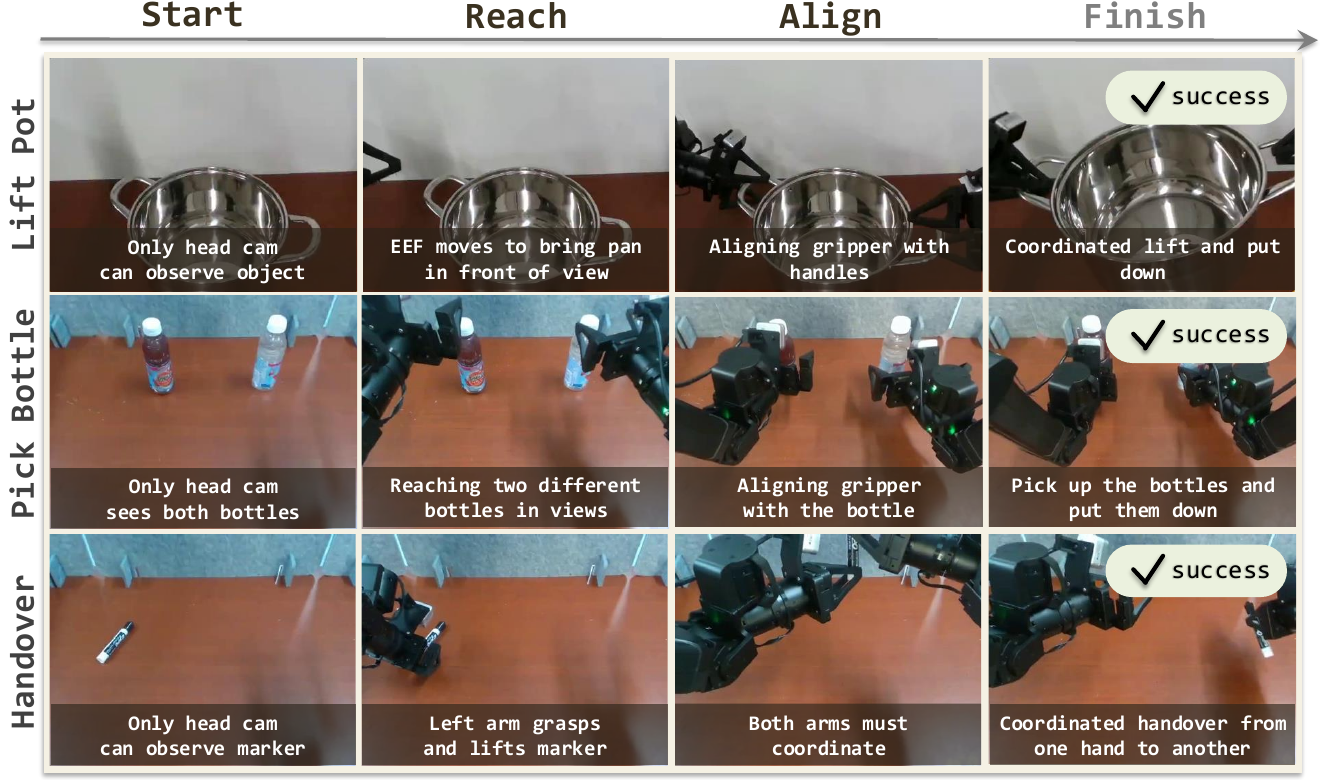}
    \vspace{-1.75em}
    \caption*{\small (a) Real-world task keyframes.}
    \includegraphics[width=\linewidth]{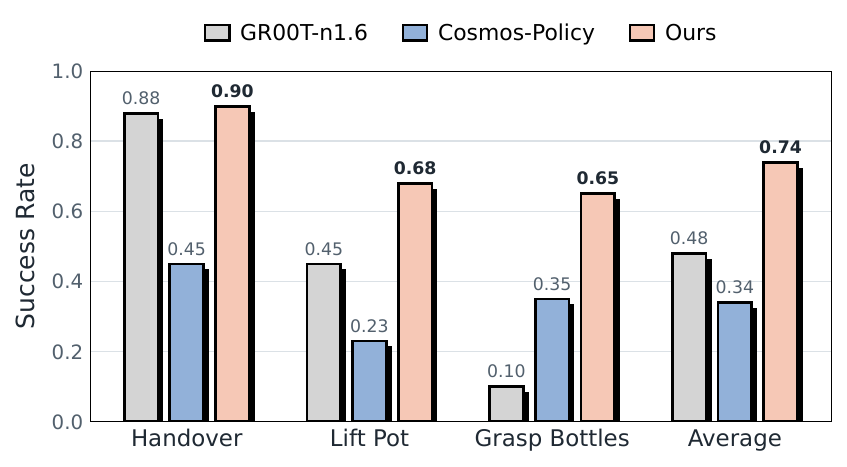}
    \vspace{-2em}
    \caption*{\small (b) Real-world success rates comparison.}
    \vspace{-0.5em}
    \caption{Real-robot keyframes for three tasks' completion and quantitative success rates comparison with the other two VLA/VAM.}
    \label{fig:real_robot_wrap}
    \vspace{-1em}
\end{wrapfigure}

\vspace{-2em}
\subsection{Real-World Deployment.} 
\vspace{-0.5em}
We evaluate \ours on three bimanual tasks using the I2RT platform: Handover Marker, Lift Pot, and Pick Up Bottles. For each task, we collect 30 expert demonstrations and fine-tune \ours, Cosmos Policy, and GR00T-N1.6 in a multi-task manner and using the same optimization budget. Each policy is evaluated over 10 trials per task. Figure~\ref{fig:real_robot_wrap}(a) presents a representative successful rollout of Lift Pot, which requires precise grasping and coordinated motion between the two arms. As shown in Fig.~\ref{fig:real_robot_wrap}(b), \ours achieves an average success rate of 74\%, compared with 48\% for GR00T-N1.6 and 34\% for Cosmos Policy. Additional detailed task-level analysis is provided in the Appendix.

\vspace{-1em}
\subsection{Learning and Data Efficiency}
\begin{figure}[t]
    \centering
    \small
    \begin{minipage}[t]{0.385\textwidth}
        \centering
        \vspace{0pt}
        \includegraphics[width=\linewidth]{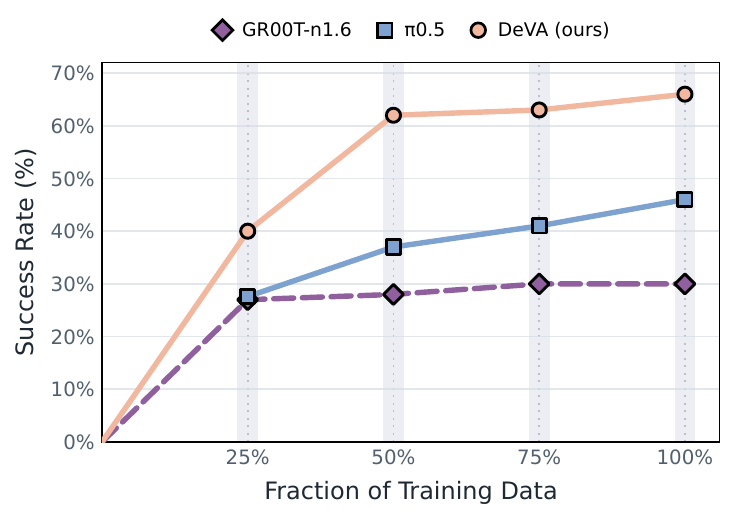}
        \vspace{-1.5em}
        \caption*{\small (a) Data percentage convergence.}
    \end{minipage}%
    \hfill
    \begin{minipage}[t]{0.385\textwidth}
        \centering
        \vspace{0pt}
        \includegraphics[width=\linewidth]{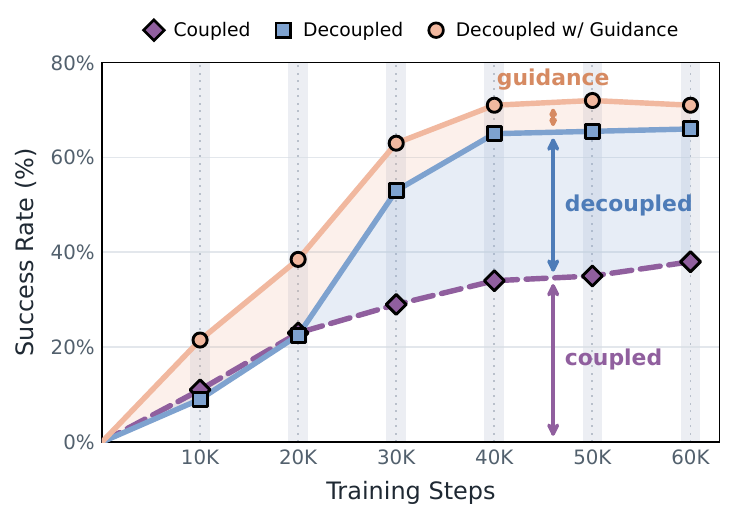}
        \vspace{-1.5em}
        \caption*{\small (b) Decoupled vs.\ Coupled convergence.}
    \end{minipage}%
    \hfill
    \begin{minipage}[t]{0.223\textwidth}
        \centering
        \vspace{0pt}
        \includegraphics[width=\linewidth]{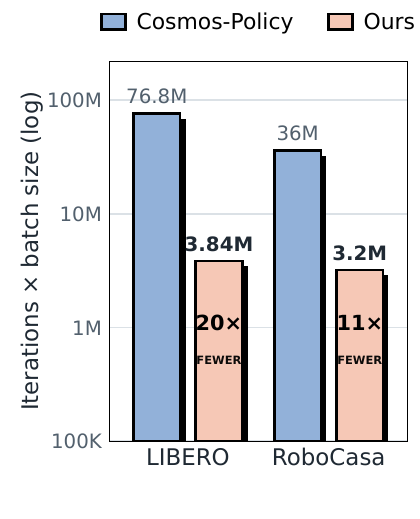}
        \vspace{-1.5em}
        \caption*{\small (c) Training efficiency.}
    \end{minipage}
    \vspace{-0.6em}
    \caption{Comparison over training optimization and data efficiency.}
    \label{fig:converge}
    \vspace{-1.8em}
\end{figure}

\vspace{-0.5em}
We evaluate training efficiency from two perspectives: optimization convergence and data budget. First, we replace the decoupled experts and their feature-transfer interface with a unified backbone while keeping the remaining training setup unchanged. As shown in Fig.~\ref{fig:converge}(b), both decoupled variants approach convergence within approximately 40-50K steps. At this point, the base and physically guided variants achieve success rates of approximately 66\% and 71\%, respectively, compared with 34\% for the unified counterpart. Based on the official reports in ~\cite{kim2025cosmospolicy}, \ours processes up to 20× fewer training examples than Cosmos-Policy as shown in Fig.~\ref{fig:converge}(c). This comparison shows that specialized experts with structured feature transfer achieve both faster convergence and higher performance under the same optimization budget.

We further train \ours and the VLA baselines using matched fractions of the RoboCasa demonstrations to reveal the learning curve. As shown in Fig.~\ref{fig:converge}(a), \ours achieves a higher success rate at every shared data budget and continues to improve as additional demonstrations are introduced. At the full data budget, \ours(base variant) reaches approximately 66\%, compared with 48\% for \(\pi_{0.5}\) and 29\% for GR00T-N1.6. Together, these results show that the decoupled architecture improves both optimization tractability and the effective use of task-specific demonstrations.

\begin{wrapfigure}{r}{0.5\textwidth}
    \vspace{-2.5em}
    \centering
    \begin{minipage}[t]{0.48\linewidth}
        \centering
        \includegraphics[width=\linewidth]{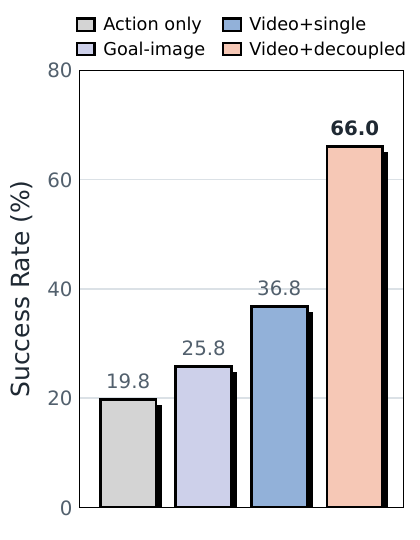}
        \vspace{-1.5em}
        \caption*{\small (a) Backbone Design.}
    \end{minipage}\hfill
    \begin{minipage}[t]{0.48\linewidth}
        \centering
        \includegraphics[width=\linewidth]{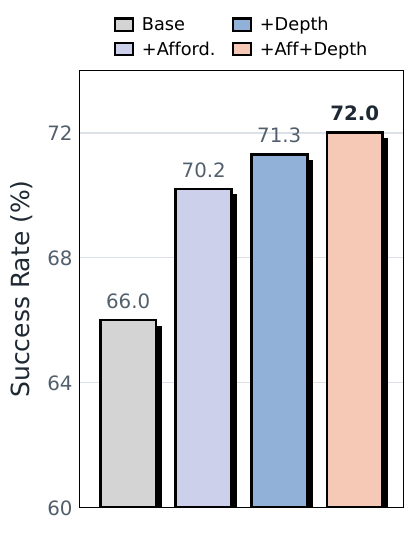}
        \vspace{-1.5em}
        \caption*{\small (b) Physical Guidance.}
    \end{minipage}
    \vspace{-0.5em}
    \caption{Ablations of the video-prediction architecture and physical-guidance components, evaluated using $\sim$55K-step checkpoints.}
    \label{fig:ablation}
    \vspace{-1em}
\end{wrapfigure}

\vspace{-1em}
\section{Ablation}
\vspace{-1em}
We ablate two aspects of DeVA on RoboCasa under a matched training budget: predictive visual modeling and physical guidance. As shown in Fig.~\ref{fig:ablation}(a), goal-image prediction improves the success rate from 19.8\% to 25.8\% over the action-only model, while future-video prediction further increases it to 36.8\%. Incorporating the decoupled architecture with multi-level feature transfer raises performance to 66.0\%, demonstrating the importance of both predictive video priors and their effective transfer to the action expert. Fig.~\ref{fig:ablation}(b) further shows that affordance and relative-depth guidance independently improve the base model. Combining both cues achieves the highest success rate of 72.0\%, indicating their complementary contributions to physically grounded action prediction.
\vspace{-1.1em}

\section{Limitation}
\vspace{-1em}
While the results are encouraging, \ours inherits some computational cost from its pretrained video backbone. Training spatiotemporal attention over future observations remains more expensive than action-only policy learning. Jointly denoising the video and action streams introduces additional inference overhead compared with models that predict only compact latent states and actions. Improving efficiency through latent-space prediction and accelerated sampling is an important direction for future work.
\vspace{-1em}

\section{Conclusion}
\vspace{-1em}
We present \ours, a decoupled Video-Action Model that transfers predictive video representations to robot control through specialized experts, multi-level feature transfer, and physically salient affordance and depth guidance. Across simulation benchmarks and real-world bimanual manipulation, \ours achieves strong performance with limited demonstrations, converges faster than a matched unified architecture, and benefits from physical guidance, demonstrating the promise of video generative priors for robot policy learning. 

\acknowledgments{This work was supported in part by funding from NSF awards \#2622839 and \#2403297. All views and conclusions expressed in this work are those of the authors and not a reflection of these
source}

\bibliography{main} 
\pagenumbering{arabic}
\renewcommand*{\thepage}{A\arabic{page}}

\appendix

\newcommand{\appendixhead}{
  \begin{center}
    {\Large \bf Appendix}
  \end{center}
  \vspace{2mm}
}

\appendixhead

\label{sup}
\section{Implementation Details}
\subsection{Video Expert Architecture}
In \ours, we adopt Cosmos-Predict2~\cite{nvidia2025cosmospredict2} as the video generative backbone and initialize it from the pretrained 480p/16-fps checkpoint. During training, the video stream is visually conditioned only on the initial observation frame, while the remaining frames serve as future prediction targets. We retain the pretrained T5 text encoder for language conditioning. Multi-view observations are spatially tiled into a single composite frame. For example, RoboCasa provides three views, agentview\_left, agentview\_right, and end\_effector\_view, which are arranged in either a $1\times3$ or $2\times2$ layout, with the unused tile in the latter left blank. We observe negligible performance differences between the two layouts. Tiling allows all views to be processed within a single video stream, reducing computation compared with processing them separately. The video backbone comprises 28 DiT blocks and approximately 2B parameters. It is trained using the standard EDM denoising objective in latent space:

\[
\mathcal{L}_{\mathrm{video}}
=
\mathbb{E}_{t}
\left[
\lambda(t)\,
\ell_{\mathrm{EDM}}\big(\hat{V}_t, V_t\big)
\right],
\]
where \(\hat{V}_t\) denotes the denoised video latent prediction at diffusion time \(t\), \(V_t\) denotes the ground-truth target latent, and \(\lambda(t)\) is the standard EDM weighting term.

\subsection{Action Expert Architecture}
The action expert follows a standard action-DiT architecture based on the action head of GR00T-N1.5~\cite{bjorck2025grootn1}. To avoid redundant visual encoding, it is conditioned directly on the intermediate video representations, T5 language embeddings, and physical guidance, without a separate visual encoder. Because the initial observation is already encoded by the video backbone, the transferred features provide both the current visual context and predictive dynamics required for action generation. An input MLP projects noisy action trajectories into the model dimension, while an output MLP maps the denoised tokens back to the action space. 

Alongside the action tokens, 12 learnable bridge tokens of dimension 1024 are provided as input and expanded across the batch to a shape of $(n,12,1024)$. Within the self-attention layers, these tokens aggregate transferred video information and propagate global context across the action stream. The action expert contains 16 layers and approximately 500M parameters, with self-attention and cross-attention interleaved across the network. Accordingly, eight layers are uniformly sampled from early to late stages of the video backbone and routed to the corresponding cross-attention layers. The action expert is trained using a flow-matching objective in the action space:
\[
\mathcal{L}_{\mathrm{act}}
=
\mathbb{E}_{t}
\left[
\left\|
v_\theta(\tilde{A}_t, t, c) - u_t
\right\|_2^2
\right],
\]

where \(t\) denotes the flow-matching timestep, \(\tilde{A}_t\) denotes the interpolated noisy action trajectory at time \(t\), \(u_t\) is the corresponding target flow, and \(c\) denotes the conditioning inputs, including observations, language instructions, and video features and physical guidance. 

\subsection{Affordance and Depth Guidance Decoder}
Intermediate features from video generative backbones provide useful perceptual representations for downstream tasks. Prior work~\cite{xu2023open} has explored transferring diffusion-model features to classification, segmentation, and other dense prediction tasks. In \ours, we attach two DPT-style decoders to the video backbone for affordance and relative-depth prediction. To capture temporal structure, temporal transformer blocks are inserted before the residual convolutional blocks, enabling information exchange across frames and encouraging temporally consistent predictions.

We uniformly sample eight intermediate layers across the video backbone to distribute physical supervision across its depth. Four layers provide multi-scale features to the affordance decoder, while the remaining four provide features to the relative-depth decoder. Because a scene may contain multiple plausible interaction regions depending on the task, the affordance decoder is conditioned on the T5 language embeddings through FiLM layers. Each decoder contains four decoding stages, and the two decoders add approximately 40M parameters in total, which is small relative to the combined video and action experts.

Features from the third stage of both decoders are further used as physical guidance for the action expert. These features are resampled to the spatiotemporal resolution of the video tokens and concatenated with the corresponding video features along the channel dimension. The fused features are then projected to the hidden dimension of the text tokens and appended to them along the token dimension. The resulting conditioning sequence provides the keys and values for the cross-attention layers in the action expert.

\subsection{Parameter Overview}
\label{sec:param-overview}
\ours combines a pretrained 2B-parameter Video2World diffusion transformer (DiT) with a specialized action expert and two lightweight physical-guidance decoders. Table~\ref{tab:param-overview} reports the parameter count of each component. The T5-XXL text encoder ($4.86$B parameters) and Cosmos continuous VAE tokenizer remain frozen throughout training, providing language-conditioning embeddings and mapping videos to and from the latent space, respectively. The components optimized across our two-stage pipeline include the Video2World DiT ($1.96$B parameters) and action expert ($564$M parameters), comprising the state/action embedders, action transformer, and cross-attention modules for video feature transfer. The affordance decoder ($17.6$M), relative-depth decoder ($13.4$M), and fusion projection ($5.2$M) add $36.3$M parameters, corresponding to only $1.4\%$ of the parameters optimized during training. The decoders are trained during Stage~1 and frozen during Stage~2, while the fusion projection injects their features into the action-conditioning pathway. Overall, \ours contains $7.55$B parameters, of which $2.56$B ($33.9\%$) are updated during the two-stage training pipeline.

\begin{table}[h]
\centering
\caption{Parameter overview of \ours with information about frozen and active modules.}
\vspace{0.5em}
\label{tab:param-overview}
\begin{tabular}{@{}lrl@{}}
\toprule
Component & Parameters & Status \\
\midrule
T5-XXL       & 4{,}864{,}791{,}552 & frozen \\
VAE tokenizer             &   126{,}892{,}531 & frozen \\
Video2world DiT           & 1{,}956{,}413{,}440 & trained \\
Action head             &   563{,}865{,}614 & trained \\
Affordance head           &    17{,}630{,}465 & trained \\
Depth head                &    13{,}432{,}065 & trained \\
Joint-fusion projection   &     5{,}244{,}928 & trained \\
\midrule
Trainable total           & 2{,}556{,}586{,}512 & \\
Frozen total              & 4{,}991{,}684{,}083 & \\
Total                     & 7{,}548{,}270{,}595 & \\
\bottomrule
\end{tabular}
\end{table}

\subsection{Training Scheme}
\label{sec:training-hparams}
Following the two-stage pipeline, we fine-tune the 2B-parameter Video2World diffusion transformer (DiT) for future-video prediction and train the action expert using action-space flow matching. Two auxiliary dense-prediction objectives supervise the affordance-heatmap and relative-depth decoders. Training is performed on eight GPUs using Fully Sharded Data Parallelism, with a per-GPU batch size of $8$ (global batch size $64$) and bfloat16 mixed precision. The optimization, data, and auxiliary-decoder settings are summarized in Tables~\ref{tab:hparams-optim}, \ref{tab:hparams-data},
and~\ref{tab:hparams-aux}, respectively.

We use FusedAdamW with a piecewise learning-rate schedule. The learning rate is linearly increased over the first $1{,}000$ steps to a peak value of $1\times10^{-4}$ and then linearly decayed to $3\times10^{-5}$ ($0.3\times$ the peak) by step $30\mathrm{k}$. It is subsequently reduced to $6\times10^{-6}$ ($0.06\times$ the peak) and held constant for the remainder of training. Gradients are clipped to a global norm of $1.0$.

\begin{table}[h]
\centering
\caption{Optimization and learning-rate schedule.}
\label{tab:hparams-optim}
\begin{tabular}{ll}
\toprule
Hyperparameter & Value \\
\midrule
Initialization & pretrained 2B video2world DiT \\
Optimizer & FusedAdamW \\
Peak learning rate & $1\times10^{-4}$ \\
Weight decay & $1\times10^{-3}$ \\
$(\beta_1,\,\beta_2)$ & $(0.9,\,0.99)$ \\
$\epsilon$ & $1\times10^{-8}$ \\
LR schedule & linear warmup $+$ linear decay \\
Warmup steps & $1{,}000$ (from $10^{-6}\!\cdot\!\mathrm{lr}$ to peak) \\
Decay & to $0.3\!\cdot\!\mathrm{lr}$ over steps $1\mathrm{k}$--$30\mathrm{k}$, then $0.06\!\cdot\!\mathrm{lr}$ \\
Gradient clipping & global norm $1.0$ \\
Training iterations & $40{,}000$ for I2RT; $\sim55{,}000$ for simulation \\
Mixed precision &  bfloat16 \\
Batch size (per-GPU / effective) & $8 \,/\, 64$ \\
Parallelism & FSDP full shard ($8$ GPUs) \\
\bottomrule
\end{tabular}
\end{table}

The three datasets differ in clip length, image resolution, and action/state representation. All actions, together with the proprioceptive states in the real-world YAM dataset, are normalized to $[-1,1]$. For real-world YAM data, the normalization bounds are computed independently for each dimension using the $1$st and $99$th percentiles. RoboCasa and LIBERO instead use min--max normalization: the first three translation dimensions use symmetric bounds defined by $\pm\max|\cdot|$, while the remaining dimensions use their empirical minima and maxima. The real-world policy is conditioned on $14$-dimensional proprioceptive states and predicts $14$-dimensional bimanual actions. The six motion dimensions of each arm are represented as deltas relative to the initial state of the clip, while the gripper dimensions remain absolute, corresponding to the mask $(6,-1,6,-1)$. RoboCasa and LIBERO predict $7$-dimensional absolute actions without proprioceptive-state conditioning. For visual input, the real-world YAM dataset horizontally tiles three synchronized views (left, head, and right). RoboCasa arranges its three views in a $2\times2$ grid with the unused fourth tile left blank, whereas LIBERO horizontally tiles its two views.

\begin{table}[h]
\centering
\footnotesize
\caption{Dataset-specific data and dataloader settings.}
\label{tab:hparams-data}
\begin{tabular}{@{}llll@{}}
\toprule
Setting & I2RT Deployment & RoboCasa & LIBERO \\
\midrule
Clip length (frames) & $25$ & $33$ & $25$ \\
Resolution ($H{\times}W$) & $128{\times}384$ & $256{\times}256$ & $128{\times}256$ \\
Camera views & $3$ (tiled) & 3 (tiled) + 1 (blank) & 2 (tiled) \\
Frame rate & $16$~fps & $16$~fps & $16$~fps \\
Action dimension & $14$ & $7$ & $7$ \\
State dimension & $14$ & -- & -- \\
Delta-action targets & yes, mask $(6,-1,6,-1)$ & no & no \\
Action/state norm. & quantile (1/99 pct.) & min-max & min-max \\
Batch size (train / val) & $8 \,/\, 4$ & $8 \,/\, 4$ & $8 \,/\, 4$ \\
\bottomrule
\end{tabular}
\end{table}

The two auxiliary decoders receive features from disjoint sets of intermediate video-DiT blocks. Each decoder constructs a four-level feature pyramid with a finest resolution of $H/4\times W/4$. Supervision is applied to latent frames at a temporal stride of $4$, yielding $1+(h-1)/4$ supervised frames from each $h$-frame clip. The eight video blocks used across both decoders map one-to-one to the eight cross-attention layers in the action expert. The affordance decoder is trained using positive-class-weighted binary cross-entropy to address foreground--background imbalance, while the relative-depth decoder combines masked $L_1$ reconstruction with an image-gradient loss for edge-aware geometric supervision. Both auxiliary losses are added to the video diffusion objective with unit weights.

\begin{table}[h]
\centering
\caption{Auxiliary affordance and relative-depth decoder settings.}
\label{tab:hparams-aux}
\begin{tabular}{
    l
    >{\centering\arraybackslash}p{0.27\linewidth}
    >{\centering\arraybackslash}p{0.27\linewidth}
}
\toprule
Hyperparameter & Affordance & Depth \\
\midrule
Loss weight & $1.0$ & $1.0$ \\
Supervised video-DiT blocks
    & $\{6,13,20,27\}$
    & $\{3,10,17,24\}$ \\

\rowcolor{sharedrow}
Pyramid resolutions
    & \multicolumn{2}{c}{$H/4{\times}W/4,\ H/8{\times}W/8,\ H/16{\times}W/16,\ H/32{\times}W/32$} \\

\rowcolor{sharedrow}
\# Supervised frames
    & \multicolumn{2}{c}{$1+\lfloor(h-1)/4\rfloor$ per $h$-frame clip} \\

\rowcolor{sharedrow}
Decoder channels
    & \multicolumn{2}{c}{$256$} \\

\rowcolor{sharedrow}
Temporal-attn.\ blocks / heads
    & \multicolumn{2}{c}{$2\,/\,8$} \\

Loss type
    & BCE with logits
    & Masked $L_1$ + masked grad. \\

Output-bias initialization
    & $\mathrm{logit}(0.05)$
    & $\mathrm{logit}(0.15)$ \\

Target channels
    & $1$ (heatmap)
    & $2$ (depth + validity) \\

\rowcolor{sharedrow}
Target resolution
    & \multicolumn{2}{c}{$H/4\times W/4$} \\
\bottomrule
\end{tabular}
\end{table}

Affordance targets are independently normalized by the maximum value in each frame, smoothed using a $3\times3$ Gaussian kernel with $\sigma{=}1.0$, and then downsampled to the decoder output resolution.

\section{Data Processing}
We train \ours on each benchmark with all the training data in a multi-task manner.
\subsection{Training Data}
\paragraph{LIBERO.} 
We train on all four LIBERO suites: LIBERO-Spatial, LIBERO-Object, LIBERO-Goal, and LIBERO-Long. Each suite contains 10 tasks with 50 demonstrations per task, yielding 2,000 demonstrations in total before filtering. Because some demonstrations do not complete successfully when replayed, we use the filtered dataset introduced by Cosmos-Policy~\cite{kim2025cosmospolicy} to improve data quality and ensure a fair comparison with prior baselines. This dataset replays each demonstration in the simulator and retains only successful trajectories.

\paragraph{RoboCasa} 
The RoboCasa training set contains 24 tasks, with more than 50 demonstrations available for each task. Following the LIBERO setup, we use the replay-filtered dataset introduced by Cosmos-Policy~\cite{kim2025cosmospolicy}, which retains successful trajectories and contains 1,199 episodes in total.

\paragraph{I2RT YAM Deployment} 
We evaluate real-world deployment on the I2RT YAM bimanual robot platform using demonstrations collected through teleoperation. The evaluation includes three tasks: (1) picking up a marker with one arm and handing it over to the other; (2) lifting a pot with both arms; and (3) picking up two bottles, one with each arm. We collect 30 demonstrations per task based on VR teleoperation~\cite{cheng2025open}, covering diverse initial object configurations. Each timestep contains three synchronized views: an external head view and two wrist-mounted end-effector views. 

\subsection{Depth Map Supervision}
To provide the depth maps for auxiliary decoder supervision, we use an offline video depth prediction model~\cite{chen2025video} to generate relative-depth targets. The predictions are reversed and normalized to $[0,1]$, such that nearby regions have values close to $0$ and distant regions have values close to $1$. To avoid boundary artifacts caused by the tiled multi-view input, we estimate depth independently for each camera view and then tile the resulting maps using the corresponding view layout. Because these targets serve as auxiliary supervision for distilling relative scene geometry into the intermediate video features, they do not require metric accuracy or exact scale consistency across views. This formulation is therefore robust to minor cross-view inconsistencies and temporal drift over long prediction horizons.

\subsection{Affordance Map Supervision}
In addition to relative depth, we apply affordance-map supervision to intermediate video features. Because oracle interaction signals are available only in simulation, we use different annotation strategies for simulated and real-world data. For simulation benchmarks, we extract contact information from the RoboSuite MuJoCo simulator and project the contact points into the image plane. For frames containing valid contacts, the mean projected contact position defines the affordance center. For frames without direct contact, we use the first valid contact location in the corresponding trajectory, representing the region that the gripper is expected to approach. The resulting locations are converted into smooth Gaussian heatmaps for supervision.

For real-world demonstrations, where oracle contact information is unavailable, we use the task-conditioned affordance model~\cite{tang2025uad} in an out-of-the-box manner to generate affordance pseudo-labels. The pretrained model distill affordance knowledge from vision and vision-language foundation models into a lightweight prediction model, allowing efficient annotation without task-specific manual labels. In practice, it provides usable affordance targets for most of our real-world manipulation scenes.

As with relative-depth supervision, affordance maps provide an auxiliary signal that encourages the video features to emphasize task-relevant interaction regions rather than serving as exact spatial targets. The supervision is therefore tolerant to moderate noise and localization errors in the generated pseudo-labels.
\begin{figure}[t]
    \centering
    \includegraphics[width=\linewidth]{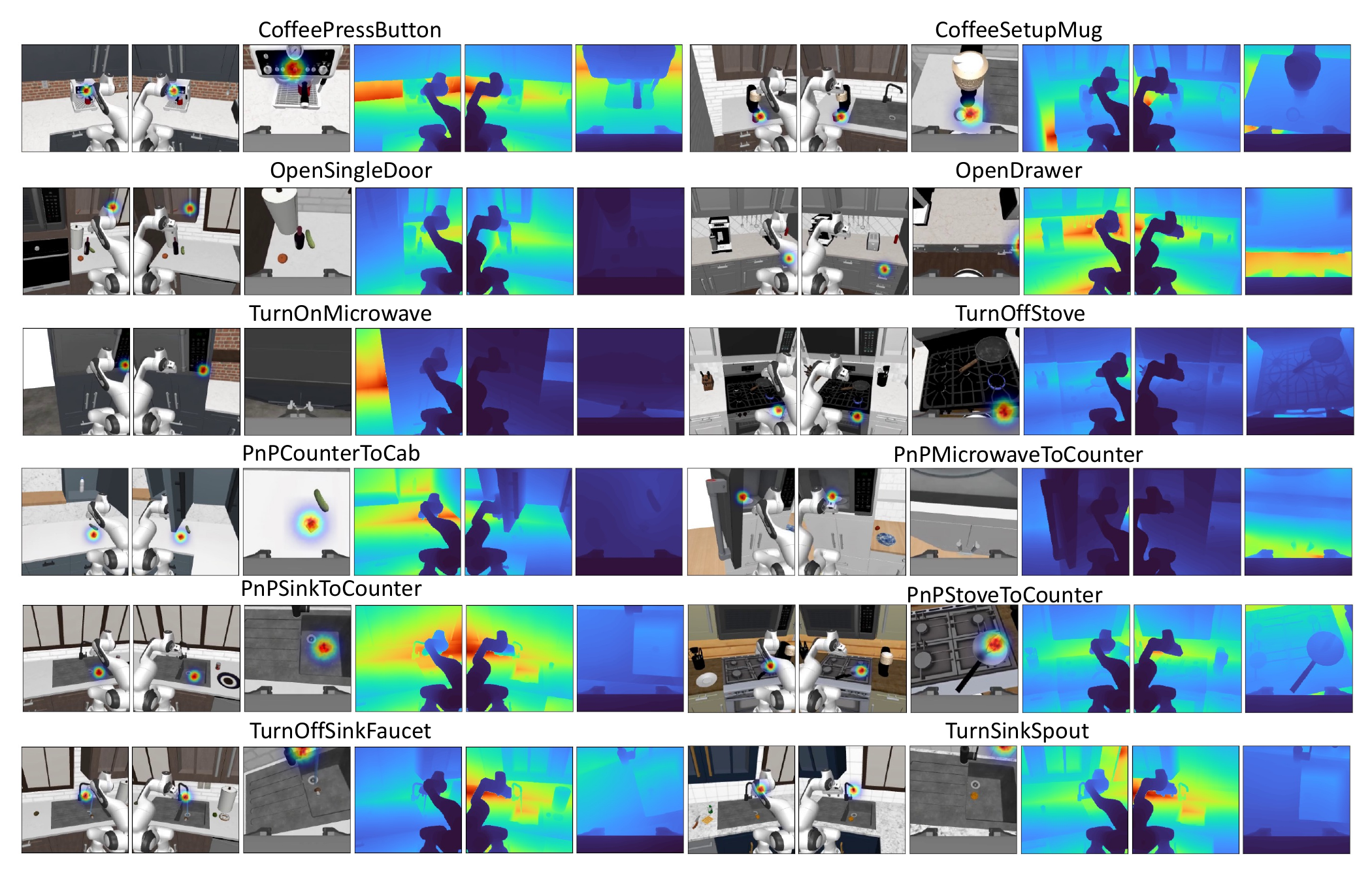}
    \vspace{-1em}
    \caption{Visualization of physical guidance across representative manipulation examples. Affordance maps highlight task-relevant interaction regions, while relative-depth maps capture the near-to-far geometric structure of each scene.}
    \vspace{-1em}
    \label{sup_fig:guidance_vis}
\end{figure}
\subsection{Guidance Visualization}
Figure~\ref{sup_fig:guidance_vis} presents representative visualizations of the physical guidance used to supervise the intermediate video-DiT features and condition the action expert. The task-conditioned affordance maps highlight interaction regions relevant to the language instruction, such as target objects and expected end-effector contact locations. The relative-depth maps provide complementary geometric information by capturing object boundaries, free space, and near-to-far scene structure. Although the supervision may contain moderate prediction noise, both signals remain spatially coherent across diverse manipulation scenes. These examples illustrate how affordance and relative depth provide complementary semantic and geometric guidance for video-to-action transfer.

\section{Real-World Deployment}
\subsection{Task Details}
\begin{figure}[h]
    \centering
\includegraphics[width=\linewidth]{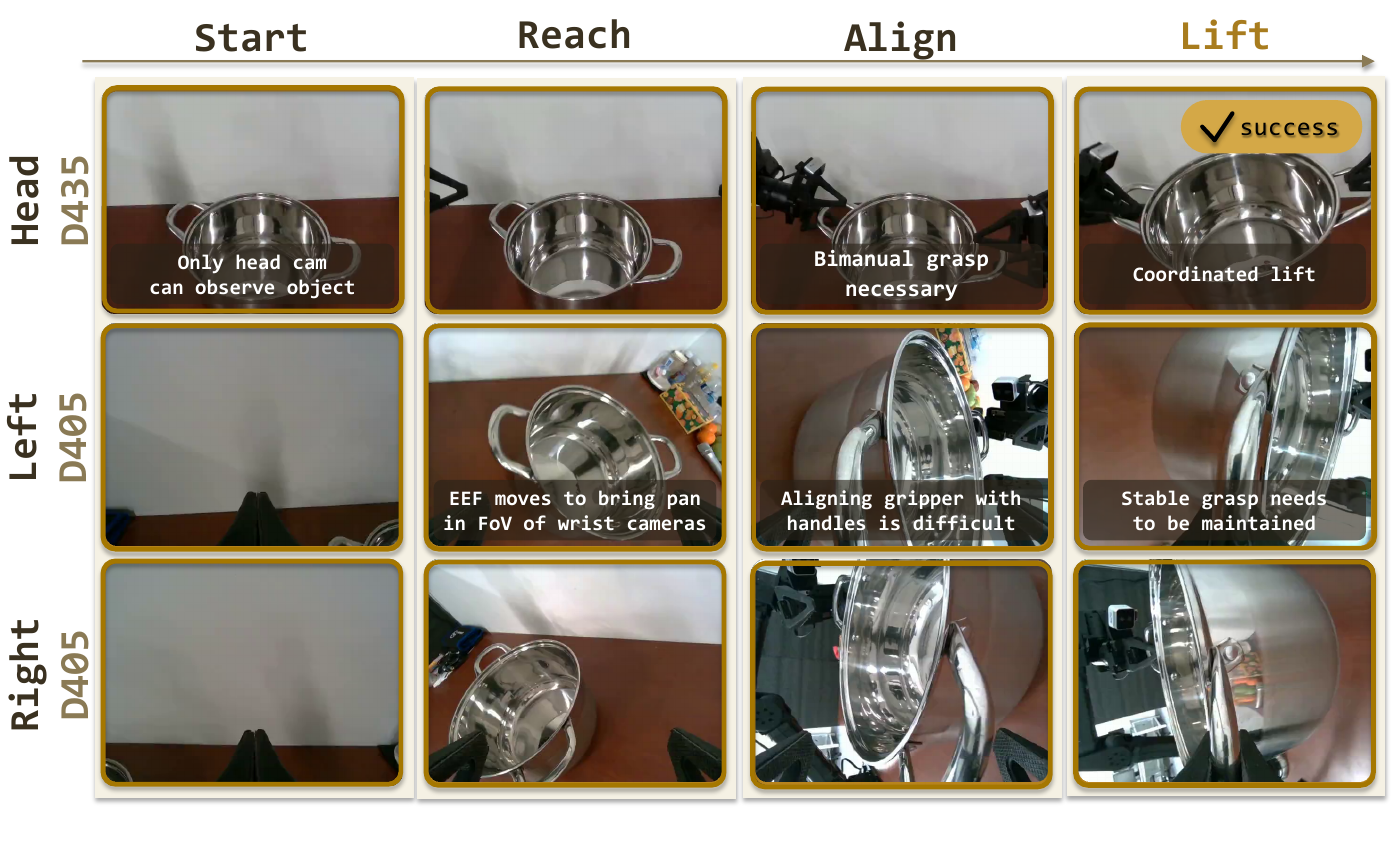}
    \vspace{-1em}
    \caption{Keyframes of the Lift Pot task, including coordinated approach, synchronized grasping, stable lifting, and placement.}
    \vspace{-1em}
    \label{sup_fig:real_lift}
\end{figure}

\begin{figure}[h]
    \centering
    \includegraphics[width=\linewidth]{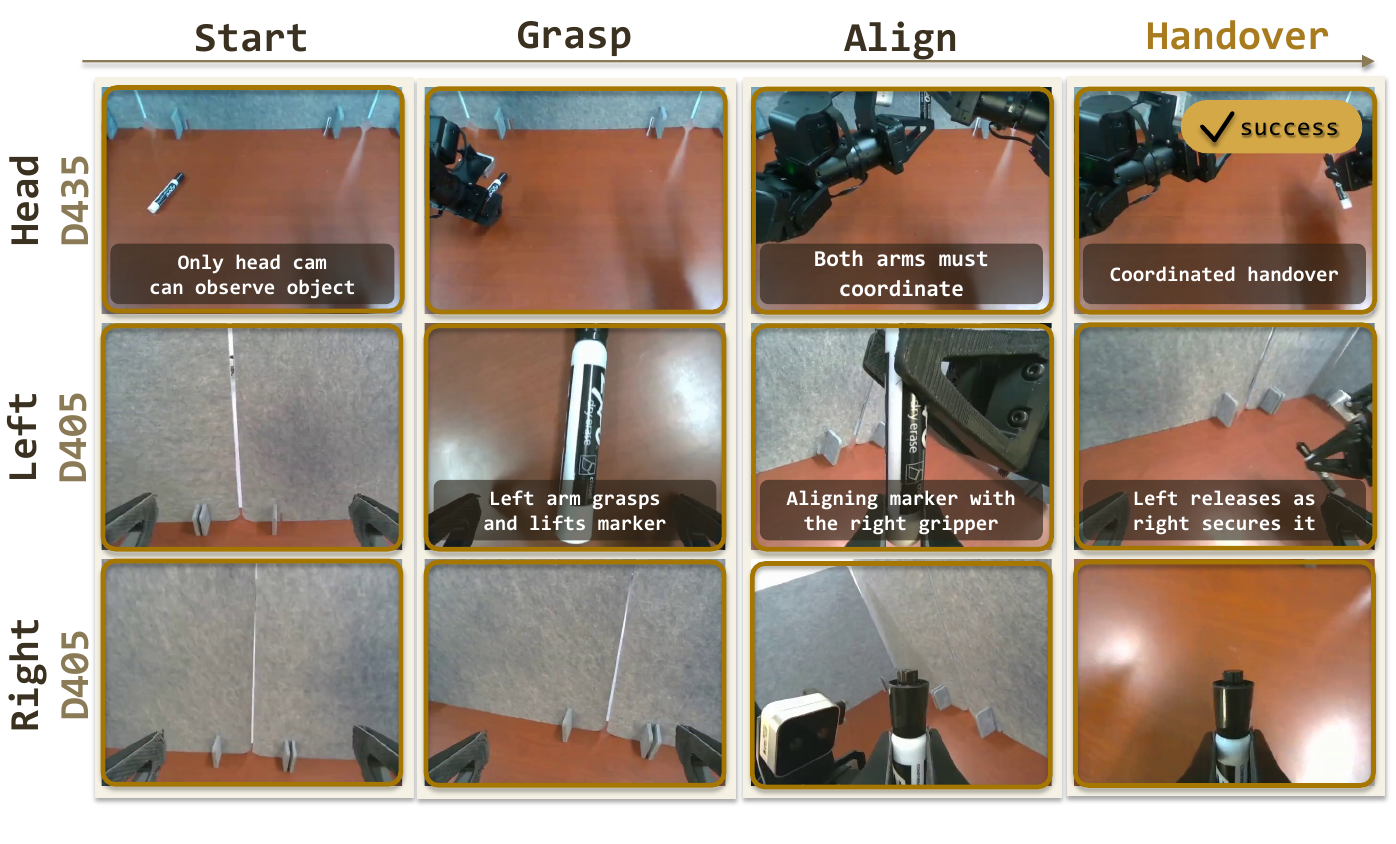}
    \vspace{-1em}
    \caption{Keyframes of the Handover Marker task, highlighting grasping, precise bimanual transfer, and stable placement.}
    \vspace{-1em}
    \label{sup_fig:real_handover}
\end{figure}

\begin{figure}[h]
    \centering
    \includegraphics[width=\linewidth]{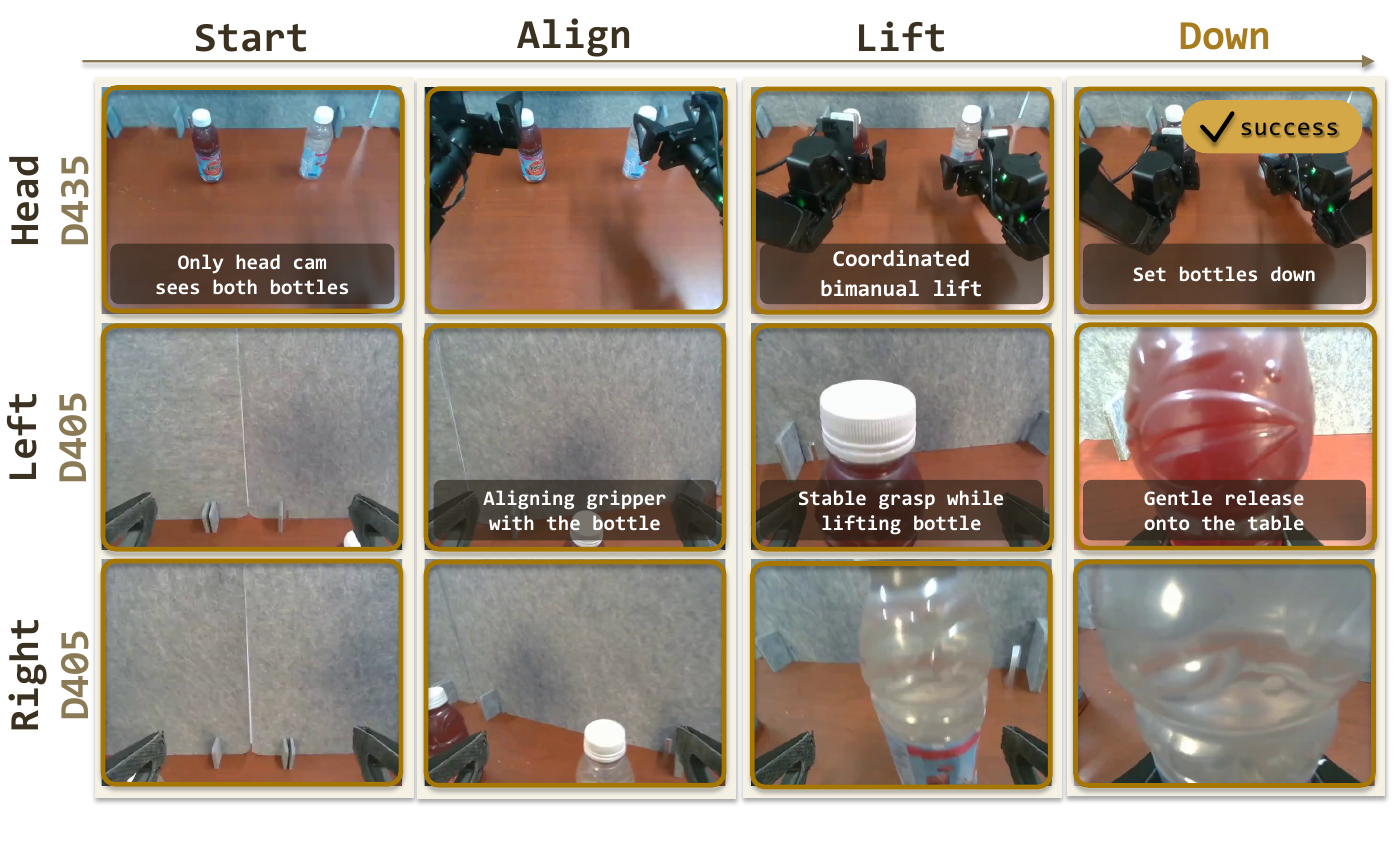}
    \vspace{-1em}
    \caption{Keyframes of the Pick Up Diverse Bottles task, including coordinated dual-arm grasping, lifting, and stable placement.}
    \vspace{-1em}
    \label{sup_fig:real_bottle}
\end{figure}

We evaluate three real-world bimanual manipulation tasks:
\begin{itemize}
    \item \textbf{Handover Marker}: A blue, black, or red marker is initially placed on either side of the workspace. The nearest arm must grasp the marker, transfer it to the opposite gripper, and place it safely on the table. The primary challenge is the handover stage, which requires precise alignment and coordinated gripper timing to prevent the marker from being dropped.
    
    \item \textbf{Lift Pot}: Both grippers approach the corresponding handle regions and align approximately parallel to the handles. They must close synchronously, lift the pot to the target height while maintaining a stable orientation, hold it securely, and return it steadily to the table. This task requires robustness to variations in pot orientation, coordination between both arms, synchronized grasping, and sufficient grip force throughout the motion. 
    
    \item \textbf{Pick Up Diverse Bottles}: The task includes four bottle instances, from which two are randomly selected and placed on the table for each episode. Each arm must approach and align its gripper with one bottle, grasp both bottles simultaneously, lift them, and place them stably back on the table. The policy must coordinate both arms across diverse bottle geometries and distinguish visually similar stages of the trajectory, particularly between lifting and placing.
\end{itemize}
Figures~\ref{sup_fig:real_lift}, \ref{sup_fig:real_handover}, and~\ref{sup_fig:real_bottle}
illustrate the keyframes required for successful completion of all three real-world tasks.

\subsection{Evaluation Score Criteria}
For more detailed analysis of rollout progress, we assign each episode a stage-completion partial credit in $[0,1]$. The score corresponds to the most advanced criterion completed during the rollout; episodes that do not satisfy the first criterion receive a score of $0$.

\paragraph{Handover Marker.}
\begin{itemize}
    \item \textbf{0.5}: One gripper lifts the marker, but the transfer to the other gripper fails.
    \item \textbf{1.0}: The receiving gripper successfully secures the marker and completes the handover.
\end{itemize}

\paragraph{Lift Pot.}
\begin{itemize}
    \item \textbf{0.25}: One gripper lifts its handle, while the other gripper fails to establish a successful grasp.
    \item \textbf{0.5}: Both grippers successfully lift the pot, but the pot is not returned to the table.
    \item \textbf{0.75}: Both grippers lift the pot, but only one gripper successfully completes the placement stage.
    \item \textbf{1.0}: Both grippers coordinate to lift the pot and return it stably to the table.
\end{itemize}

\paragraph{Pick Up Diverse Bottles.}
\begin{itemize}
    \item \textbf{0.25}: One gripper lifts one bottle, but does not return it to the table.
    \item \textbf{0.5}: Both bottles are lifted but not returned to the table, or one arm completes the full pick-and-place sequence while the other arm remains inactive.
    \item \textbf{0.75}: Both bottles are successfully lifted, but only one is returned stably to the table.
    \item \textbf{1.0}: Both arms successfully lift their assigned bottles and return them stably to the table.
\end{itemize}

\subsection{Additional Baseline Comparisons}
\paragraph{Baselines}
\ours is compared with Cosmos-Policy, the most closely related Video-Action Model, and two other pretrained VLA baselines after task-specific finetuning:
\begin{itemize}
    \item \textbf{Cosmos-Policy}: As the most closely related Video-Action Model, Cosmos-Policy adapts a pretrained Cosmos video generative backbone for robot control. We initialize it from the pretrained video checkpoint and train the action modality from scratch on the same task demonstrations used for all real-world comparisons. The model is trained for $40\mathrm{k}$ steps using its official training pipeline and configuration, without modifying its architecture or objective.
    
    \item \textbf{$\pi_{0.5}$}: We initialize $\pi_{0.5}$ from its released pretrained checkpoint and fine-tune it on the same task demonstrations used for Cosmos-Policy and \ours. The model is trained for $40\mathrm{k}$ steps using the official training pipeline and model-specific configuration, without modifying its architecture or training objective. The pretrained model has been pretrained on a large and diverse corpus of robot demonstrations.

    \item \textbf{GR00T-N1.6}: We similarly initialize GR00T-N1.6 from its released pretrained checkpoint and fine-tune it on the same task demonstrations for $40\mathrm{k}$ steps. We retain the official training pipeline and configuration provided with the model.
\end{itemize}

This setup matches the training data and number of optimization steps across methods while preserving the model-specific training procedures recommended by their respective authors.

\paragraph{Analysis}
We evaluate \ours and the three baselines on all three tasks, using 10 rollouts for each method--task pair. The quantitative results are reported in Table~\ref{tab:sub_real_robot}. Qualitative inspection further reveals several recurring failure modes.

\begin{table}[t]
\centering
\scriptsize
\setlength{\tabcolsep}{4pt}
\renewcommand{\arraystretch}{1.05}
\resizebox{0.8\textwidth}{!}{%
\begin{tabular}{lcccc}
\toprule
\textbf{Success Rate} & \textbf{Handover Marker} & \textbf{Lift Pot} & \textbf{Pick Up Bottles} & \textbf{Average} \\
\midrule
GR00T-n1.6~\cite{bjorck2025grootn1} & \underline{0.88} & 0.45 & 0.10 & 0.48 \\
$\pi_{0.5}$~\cite{intelligence2025pi0.5} & 0.70 & \underline{0.55} & \textbf{0.75} & \underline{0.67} \\
Cosmos-Policy~\cite{kim2025cosmospolicy} & 0.45 & 0.23 & 0.35 & 0.34 \\
\textbf{Ours} & \textbf{0.90} & \textbf{0.68} & \underline{0.65} & \textbf{0.74}\\
\bottomrule
\end{tabular}%
}
\vspace{1em}
\caption{Real-world deployment comparison.}
\vspace{-2em}
\label{tab:sub_real_robot}
\end{table}

\noindent\textbf{VLA models}: 
The two VLA baselines exhibit different behaviors across tasks. Notably, $\pi_{0.5}$ is initialized from a checkpoint with extensive robot-centric pretraining on large-scale and diverse robot trajectories, providing strong manipulation priors before task-specific fine-tuning. It generalizes well to variations in initial object placement, but still encounters coordination failures. During Handover Marker, the transferring gripper sometimes releases the marker before the receiving gripper establishes a stable grasp, causing the marker to fall. During Lift Pot, performance is sensitive to changes in the pot and handle orientation relative to the demonstrations. In some rollouts, the policy successfully lifts the pot but fails to initiate the placement stage, leaving it suspended. GR00T-N1.6 exhibits similar pose sensitivity on Lift Pot. On Pick Up Diverse Bottles, it frequently approaches the bottles but does not close the grippers or initiate lifting. One possible explanation is ambiguity between visually similar stages before grasping and after placement, which may cause premature task completion.

\noindent\textbf{VAM models}:
Under the same $40\mathrm{K}$-step fine-tuning budget, Cosmos-Policy adapts more slowly than all others and remains less robust to variations in object placement. During Handover Marker, it often reaches toward a fixed workspace region rather than adapting precisely to the marker location. Similar sensitivity to object pose is observed in the pot and bottle tasks. In contrast, \ours more consistently adjusts its trajectories to these variations. 

Among all methods, $\pi_{0.5}$ performs particularly well on Pick Up Diverse Bottles. We hypothesize that this advantage is partly attributable to its extensive robot-centric pretraining and its longer action horizon of $50$ steps, compared with $25$ steps for the other methods, which may help maintain coherent actions across the grasp, lift, and placement stages.

\begin{figure}[h]
    \centering

    \begin{subfigure}{0.86\linewidth}
        \centering
        \includegraphics[width=\linewidth]{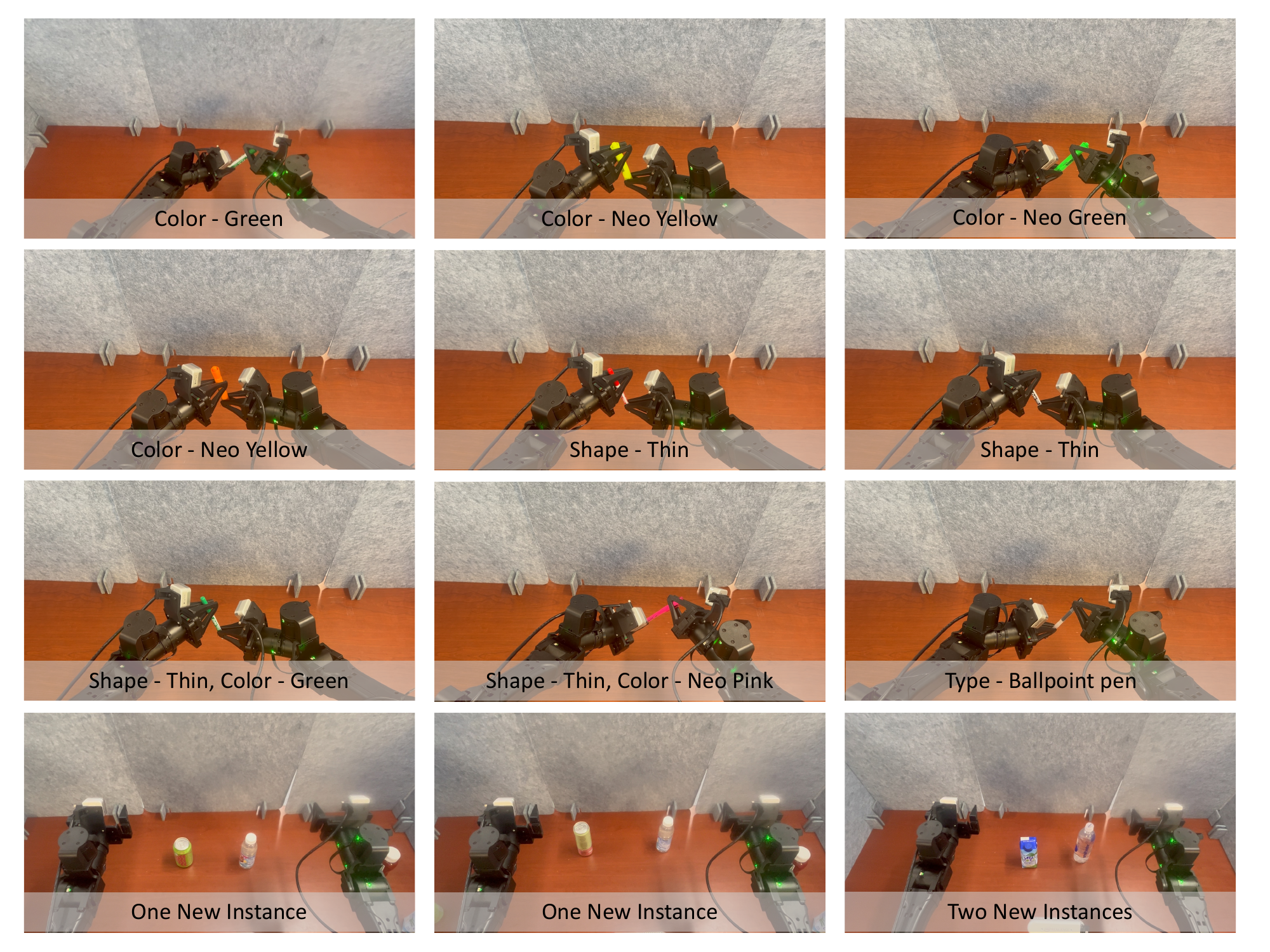}
        \caption{Instance axis.}
        \label{sup_fig:robust_instance}
    \end{subfigure}

    \vspace{0.5em}

    \begin{subfigure}{0.86\linewidth}
        \centering
        \includegraphics[width=\linewidth]{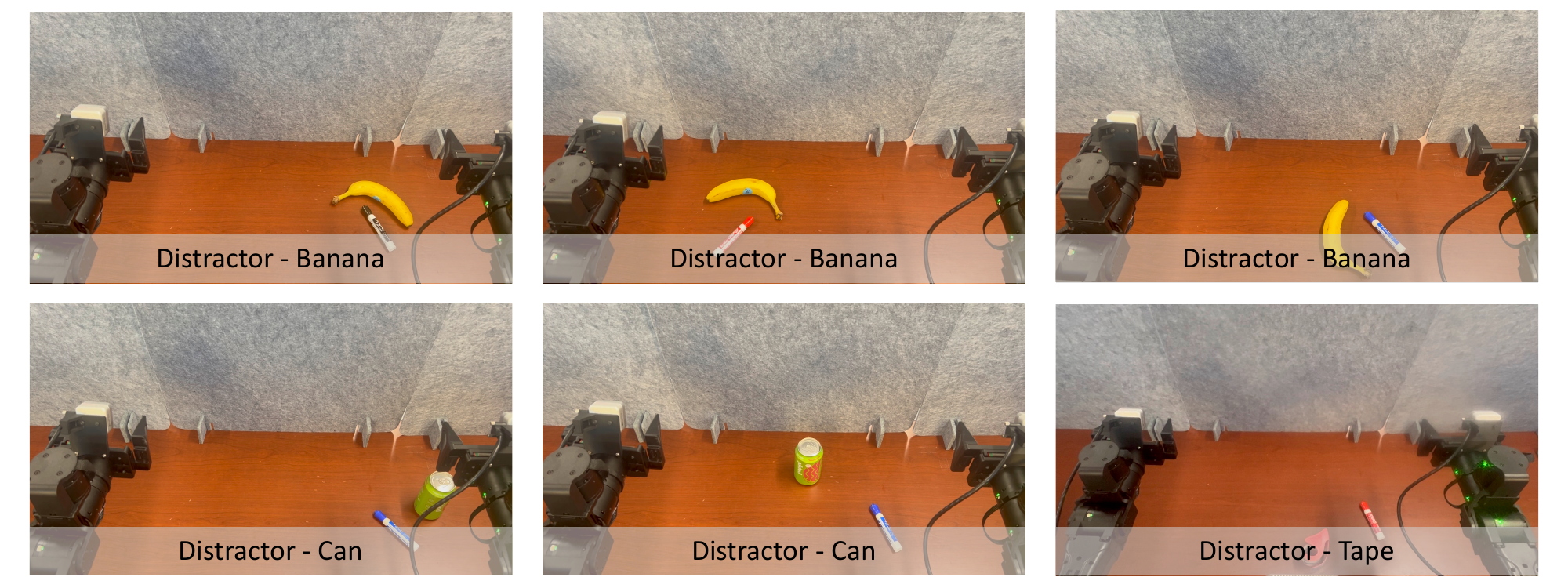}
        \caption{Distraction axis.}
        \label{sup_fig:robust_distractor}
    \end{subfigure}

    \vspace{0.5em}

    \begin{subfigure}{0.86\linewidth}
        \centering
        \includegraphics[width=\linewidth]{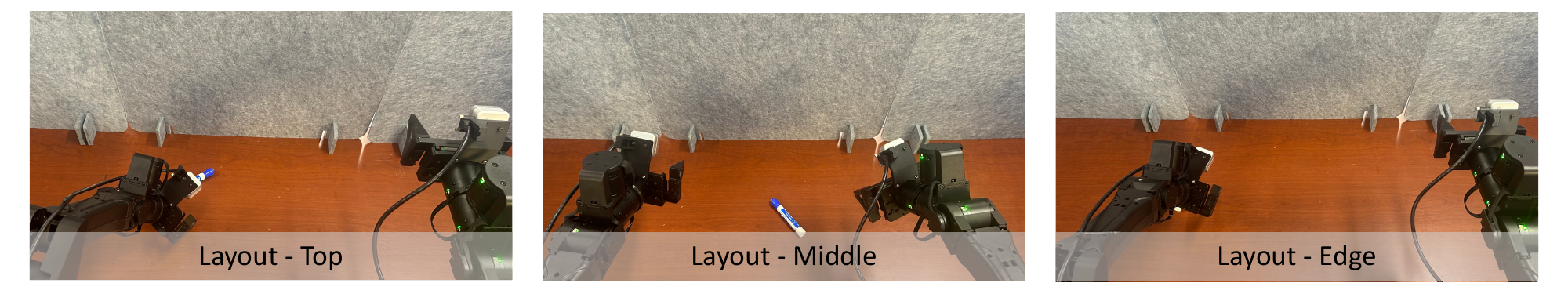}
        \caption{Layout axis.}
        \label{sup_fig:robust_layout}
    \end{subfigure}

    \vspace{-0.5em}
    \caption{Robust analysis examples along three axes: instance, distraction, and layout.}
    \label{sup_fig:robust_all}
    \vspace{-1em}
\end{figure}

\subsection{Robustness Analysis for \ours}
We conduct qualitative stress tests along three axes: (1) unseen object instances, (2) distractors and occlusions, and (3) out-of-distribution layouts. We use Handover Marker as a representative task. For unseen instances, we progressively vary the target object's color, thickness, and category, including shifts from markers to ballpoint pens and bottles. For distractor evaluation, we place a similarly elongated object, such as a banana, near the target marker or introduce a can that partially occludes the target from different camera views. Finally, we evaluate challenging layouts with object placements substantially different from those observed during training. Representative examples are shown in Figures~\ref{sup_fig:robust_instance}, \ref{sup_fig:robust_distractor}, and~\ref{sup_fig:robust_layout}.

Qualitatively, \ours remains effective under changes in object appearance and retains successful behavior for several category-level variations. It generally identifies the instructed target when distractors are spatially separated and remains robust in more challenging cases involving adjacent, shape-similar, or partially occluding objects. The model also completes several tasks under substantially shifted object layouts. Notably, these behaviors emerge from only 30 task-specific demonstrations without large-scale robot-action pretraining, suggesting that \ours can adapt its pretrained video representations to previously unseen real-world configurations.

\section{More Ablation Study}
We report additional ablations conducted during model development, focusing on the feature-interaction mechanism, observation-conditioning sources, and the use of representations from multiple video denoising steps. These studies complement the component ablations presented in the main paper.

\subsection{Feature-Interaction Mechanism}
We compare several mechanisms for transferring video representations to the action expert, including our cross-attention with bridge tokens, self-attention-based fusion, and cross-attention augmented with additional transformation layers. As shown in Table~\ref{sup_table:ablate_attention}, most direct interaction mechanisms achieve comparable performance, indicating that predictive video representations provide useful conditioning through multiple transfer interfaces. However, introducing additional transformation layers degrades performance, suggesting that excessive processing may weaken the information already encoded by the pretrained video backbone. We therefore adopt cross-attention with bridge tokens as a compact interface that supports both layer-wise feature transfer and global information aggregation. 

\begin{table}[h]
\centering
\scriptsize
\setlength{\tabcolsep}{4pt}
\renewcommand{\arraystretch}{1.05}
\resizebox{0.8\textwidth}{!}{%
\begin{tabular}{lc}
\toprule
\textbf{Model} & \textbf{Success Rate (\%)} \\
\midrule
Decoupled Branches with direct self-attention in action head & 66.79\\
Decoupled Branches with direct cross-attention & 66.50 \\
Decoupled Branches with specific layers for each cross attention & 63.75\\
Decoupled Branches with aggregated attention before cross attention & 62.75\\
\bottomrule
\end{tabular}%
}
\vspace{1em}
\caption{Ablation of video-to-action interaction mechanisms on RoboCasa. Direct self-attention and cross-attention perform similarly, while additional transformation or aggregation layers reduce performance.}
\vspace{-2em}
\label{sup_table:ablate_attention}
\end{table}

\subsection{Conditioning Sources}
We further examine whether the action expert benefits from a separately encoded initial observation in addition to the multi-level video features and language embeddings. This ablation is conducted on RoboCasa using the base model. As reported in Table~\ref{sup_table:ablate_cond}, explicitly adding the initial-frame features does not provide a consistent improvement. Because the video backbone is already conditioned on the initial observation, its intermediate representations retain the corresponding visual context. A separate initial-frame pathway therefore introduces largely redundant information.
\begin{table}[h]
\centering
\scriptsize
\setlength{\tabcolsep}{4pt}
\renewcommand{\arraystretch}{1.05}
\resizebox{0.8\textwidth}{!}{%
\begin{tabular}{lc}
\toprule
\textbf{Model} & \textbf{Success Rate (\%)} \\
\midrule
Decoupled Branches (T5 + Initial Observation + Video for Action) & 66.75\\
Decoupled Branches (Only T5 + Video for Action) & 66.50 \\
\bottomrule
\end{tabular}%
}
\vspace{1em}
\caption{Ablation of conditioning sources for the action expert on RoboCasa. Adding a separately encoded initial observation yields only a marginal $0.25$-point improvement, suggesting that its visual information is already captured by the video features.}
\vspace{-2em}
\label{sup_table:ablate_cond}
\end{table}
\subsection{Denoising-Step Conditioning}
We additionally compare conditioning the action expert on features from only the final video denoising step with aggregating features across multiple denoising steps. Although the final-step representation corresponds to the least corrupted video latent, it may be more specialized toward video reconstruction and does not necessarily provide the most informative representation for action prediction. As shown in Table ~\ref{sup_table:ablate_step_cond}, final-step conditioning achieves a success rate of $49.6\%$, while multi-step conditioning improves it to $53.2\%$. This $3.6$-point gain suggests that intermediate denoising representations provide complementary information to the action expert. Moreover, freezing the video backbone under final-step conditioning reduces performance to $22.4\%$, demonstrating that joint adaptation of the video representations is more important than simply extracting features from the pretrained backbone.

\begin{table}[h]
\centering
\scriptsize
\setlength{\tabcolsep}{4pt}
\renewcommand{\arraystretch}{1.05}
\resizebox{0.65\textwidth}{!}{%
\begin{tabular}{lc}
\toprule
\textbf{Video Conditioning Strategy} & \textbf{Success Rate (\%)} \\
\midrule
Final denoising step
    & 49.6 \\
Final denoising step + frozen video backbone
    & 22.4 \\
Multi-step denoising features
    & \textbf{53.2} \\
\bottomrule
\end{tabular}%
}
\vspace{1em}
\caption{Ablation of video denoising-step conditioning on RoboCasa. Multi-step conditioning improves over final-step conditioning by $3.6$ points, while freezing the video backbone substantially degrades performance. All results are obtained from 30K-step checkpoints.}
\label{sup_table:ablate_step_cond}
\end{table}
\vspace{-2em}
\section{Reproducibility Statement}
\vspace{-1em}
To facilitate reproducibility, we release the training and evaluation code, model configurations, and benchmark scripts. We provide detailed implementation specifications covering data preprocessing, action representations, model architecture, training schedules, hyperparameters, and pretrained checkpoint initialization. For simulation experiments, we follow the standard protocols of LIBERO, LIBERO-Plus, and RoboCasa and report results using consistent numbers of evaluation trials and random seeds when applicable. For real-world experiments, we document the robot platform, task definitions, observation modalities, control frequency, and evaluation criteria.

\end{document}